\documentclass[journal]{IEEEtran}
\usepackage{textcomp}

\usepackage{cite}

\ifCLASSINFOpdf
\else
\fi
\usepackage[]{graphicx}

\usepackage[cmex10]{amsmath}
\usepackage[tight,footnotesize]{subfigure}
\begin{document}
%
\title{A Flexible and Robust Large Scale Capacitive Tactile System for Robots}
%
%
%

\author{Perla~Maiolino,
        Marco~Maggiali,
        Giorgio~Cannata,
        Giorgio~Metta,
        and~Lorenzo~Natale
\thanks{P. Maiolino, G. Cannata are with the Department of Informatics, bioengineering, robotics and system engineering, Genova, Italy e-mail: (perla.maiolino@unige.it, giorgio.cannata@unige.it).}
\thanks{M. Maggiali, L. Natale, G. Metta are with the Istituto Italiano di Technologia (IIT), iCub Facility, Via Morego 30, Genova, Italy e-mail: (marco.maggiali@iit.it, giorgio.metta@iit.it, lorenzo.natale@iit.it).}
\thanks{Manuscript received December 14, 2012; The research leading to these results has received funding from the European Communitys Seventh Framework Programme (FP7/20072013) under grant agreement N$^{\textrm{o}}$ 231500 (Project ROBOSKIN) and grant agreement N$^{\textrm{o}}$ 288553 (Project CLOPEMA).}
\thanks{''Copyright \copyright 2013 IEEE, published in IEEE Sensor Journal, Vol. 13, Issue 10, 2013, DOI: 10.1109/JSEN.2013.2258149."}}
\maketitle

\begin{abstract}

Capacitive technology allows building sensors that are small, compact and have high sensitivity. For this reason it has been widely adopted in robotics. In a previous work we presented a compliant skin system based on capacitive technology consisting of triangular modules interconnected to form a system of sensors that can be deployed on non-flat surfaces. This solution has been successfully adopted to cover various humanoid robots. The main limitation of this and all the approaches based on capacitive technology is that they require to embed a deformable dielectric layer (usually made using an elastomer) covered by a conductive layer. This complicates the production process considerably, introduces hysteresis and limits the durability of the sensors due to ageing and mechanical stress.

In this paper we describe a novel solution in which the dielectric is made using a thin layer of 3D fabric which is glued to conductive and protective layers using techniques adopted in the clothing industry. As such, the sensor is easier to produce and has better mechanical properties. Furthermore, the sensor proposed in this paper embeds transducers for thermal compensation of the pressure measurements. We report experimental analysis that demonstrates that the sensor has good properties in terms of sensitivity and resolution. Remarkably we show that the sensor has very low hysteresis and effectively allows compensating drifts due to temperature variations.

\end{abstract}


%
\IEEEpeerreviewmaketitle

\section{Introduction}

\IEEEPARstart{T}{actile} sensing is a key technology for new generations of robots that can operate in unstructured environment and in close interaction with humans. 
Several solutions have been proposed in the literature and different transduction methods have been used with the final goal to give robots the sense of touch (see~\cite{Dahiya10} for a review). In particular the piezoresistive approach has been used massively for MEMS,  silicone based capacitive tactile sensors have been adopted for commercial sensors~\cite{Muhammad2011} and MEMS, while optical transduction and piezoelectric sensors have been used for dynamic tactile sensing~\cite{kadowaki2009, choi2005, tajima2002}. 

The first requirement for tactile sensors in the robotic domain is the resolution.  In this respect there have been promising results in the field of microelectronics leading to high density tactile sensors~\cite{shear2007,cmos2010,mems2011,mannsfeld2010}. Effective integration of tactile sensors on real robots however is still beyond the state of the art of these technologies since it requires flexible sensors that can be deployed on curved surfaces and solving system-level issues like wiring, networking, power consumption as well as cost of production, integration and maintenance.

Among the available transduction methods capacitive sensing has been widely adopted with good sensitivity and resolution~\cite{Gray96,Schmidt06,Miyazaki84}. The major drawback of this technology is that it requires to deploy deformable dielectric and conductive layers. The typical solution is to adopt elastomers, which, however, have in general a not-linear mechanical response and may introduce severe hysteresis and creep. Another problem with capacitive sensors is that they are sensitive to electromagnetic interference and to thermal variations.

In previous work we proposed a modular system for large areas tactile sensors using capacitive technology~\cite{Schmitz08, Schmitz2011}. This system was successfully employed to cover large areas of different robots with a considerable number of tactile units (approximately 2000 sensing units in the iCub, developed at Istituto Italiano di Tecnologia, and Schunk LWA arm, 1500 on the WAM arm from Barret Technology and 200 units on the Nao from Aldebaran Robotics). This first version of the tactile system demonstrated good performance in terms of sensitivity and resolution~\cite{Schmitz2011}, but it had also the following problems: i) hysteresis due to the silicone foam used to make the dielectric layer ii) reduced sensitivity due to ageing of the elastomer and poor mechanical strength to wear and tear of the external conductive layer and iii) drift due to temperature variations. 

In this paper we present a revised version of the skin system designed to overcome these limitations. In this new version the dielectric layer consists of a deformable 3D fabric on top of which are glued a conductive and a protective layer. We performed an extensive experimental characterisation of the sensor. Results show that the sensor has very low hysteresis, high sensitivity and resolution.  We furthermore embedded in the system capacitors that are insensitive to pressure and can be used for temperature compensation. We show that the output of these \emph{thermal} units can be used effectively to compensate drifts due to temperature variations.

The remainder of the paper is organised as follows. Section~\ref{sec:state-of-the-art} illustrates the state of the art in the field of tactile sensing. In Section~\ref{sec:roboskin} we describe the tactile system in detail and the improvements with respect to the previous version. Section~\ref{sec:exp} reports the experimental tests we performed. Finally in Section~\ref{sec:conclusion} we discuss the conclusions.

\section{State of the Art}
\label{sec:state-of-the-art}

Example of large scale tactile systems have been proposed in the literature. Inaba et al.\cite{Inaba96} describe a tactile sensor system composed of a layered structure of electrically conductive fabric implementing a matrix of pressure sensitive switches. Iwata et al. \cite{Iwata09} presented force detectable surface covers where the information originating from both resistive and force sensors are used to correlate pressure information and exerted force.
Ohmura et al.~\cite{Ohmura06} proposed a conformable and truly scalable robot skin system formed by self-contained modules that can be interconnected; each module, made of flexible printed circuit boards (FPCBs), contains 32 tactile elements consisting of a photo-reflector covered by urethane foam. In order to adjust the distance between each tactile sensor element, a band-like bendable substrate that can be easily folded (or even cut) is adopted. Mukay et al. \cite{Mukai08} have developed a tactile sensor system for the robot RI-MAN which uses FPCBs with a tree-like shape to conform to curved surfaces and commercially available piezoresistive semiconductors as pressure sensors. The tactile sensor systems developed for the robot ARMAR-III by Asfour~\cite{Asfour06}, uses skin patches, specifically designed for the different parts of the robot body that have flat or cylindrical shape and that are based on piezoresistive sensor matrices with embedded multiplexers.
Tactile sensing for the robot KOTARO \cite{Mizuuchi06} was achieved by using a layer of pressure sensitive conductive rubber, sandwiched by flexible bandages formed by two FPCBs that incorporate 64 taxels each. Minato et al. proposed piezoelectric transducers for the humanoid robots CB2 \cite{Minato07}.
Shimojo et al. \cite{Shimojo04} developed a mesh of tactile sensors arranged as a net where only nearby taxels are connected through wires; the tactile sensor mesh is able to cover surfaces of arbitrary curvatures, but the shape of a patch must be specifically designed. 
Mittendorfer et al. \cite{Mitt11} developed a tactile sensor system made by small hexagonal PCB modules equipped with multiple discrete off-the-shelf sensors for temperature, acceleration and proximity. Each module contains a local controller that pre-processes the sensory signals and actively routes data through a network of modules towards the closest PC connection. The sensory system is embedded into a rapid prototyped elastomer skin material and
redundantly connected to neighbours modules by four ports.
his solution is modular and conformable to the shape of the robot body surfaces, the electronics is embedded and the computation distributed; the soft dielectric layer allows compliance ensuring safe interaction with human and environment; the system has been developed using COTS components allowing a decreasing of the manufacturing cost, and furthermore has been easily integrated on several humanoid and industrial robots.

\section{Robot skin tactile system}
\label{sec:roboskin}

The tactile system described in this paper is based on the one presented in~\cite{Schmitz2011}. The basic transduction mechanism is implemented as a capacitor in which the dielectric deforms when pressure is applied. The basis of the sensor is a flexible FPCB in which a conductive area form the first plate of the capacitor. On top of the FPCB there is a deformable dielectric and a conductive layer which provides the second plate of the capacitor and works as a common ground plane that protect the sensor from electromagnetic interferences. The FPCB is shaped as a triangle hosting 12 sensors (i.e., taxels) and a capacitance to digital converter (CDC, AD7147 from Analog Devices) that measures the capacitance of each sensor, performs the AD conversion and transmits the values to a serial line. Several triangles can be interconnected to form a mesh of sensors to cover the desired area. The triangles and the connections among them are flexible: the resulting mesh can therefore be adapted to curved surfaces (details can be found in~\cite{Schmitz2011}). 
The CDC can measure the capacitance of all taxels with 16 bit resolution, although the actual dynamic range of the sensor implementation has 8 bit resolution. The CDC has an $I^{2}C$ bus so that up to 4 modules can communicate over the same serial line; three communication ports placed along the sides of the triangle relay the signals from one triangle to the adjacent ones. Up to 16 triangles can be connected in this way (4 serial buses with 4 different addresses each) and only one of them needs to be connected to a microcontroller board (i.e., MTB), which sends the measurements over a CAN bus to the PC. 
The measurements sent from the CDC chip are the result of an averaging process. The CDC output can be delivered at different frequencies, but in our current implementation it is delivered at 25 Hz.

\begin{figure}[!t]
\centering
\includegraphics[width=\columnwidth]{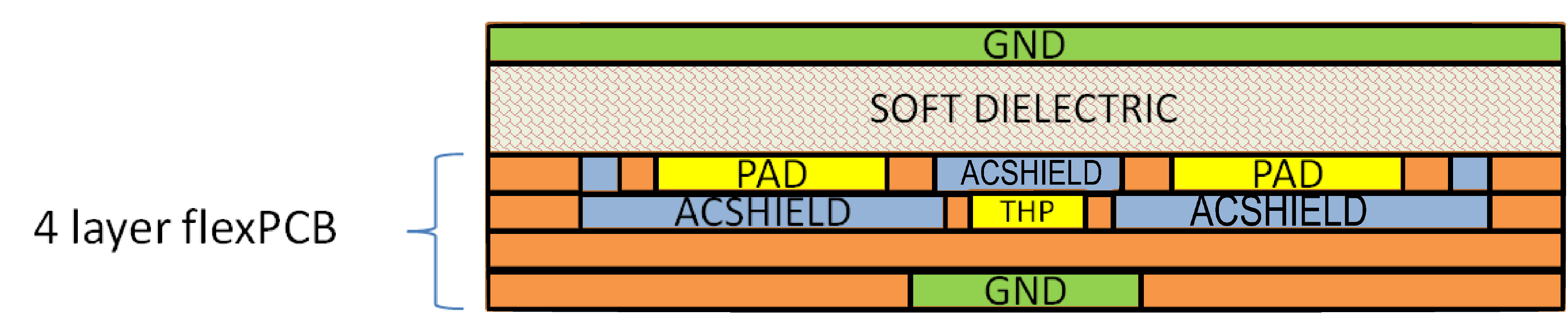}
 \caption{In the Figure the vertical section of the structure of the new tactile module. The flexible FPCB consists of four layers. The pads in yellow represent one of the plates of two capacitors that are sensitive to pressure, while the layer on top (GND) is made with conductive material and forms the second plate of these capacitors. Pressure variations deform the soft dielectric and affect the value of capacitance. We embedded in the FPCB special pads (THP) that form capacitors that are insensitive to pressure. The output of the thermal pads is affected only by temperature and can be used to compensate the thermal drift of the other sensors in the module.}
\label{struttura}
\end{figure}

One of the problem with this system was that it had a relevant drift due to temperature variations. We empirically determined that temperature variations were induced by deformations in the FPCB and not in the dielectric. This inspired us to develop a revision of the FPCB in which two of the 12 taxels have been \emph{embedded} within the FPCB (see Figure \ref{struttura}). Since the FPCB is bonded to a rigid support surface, pressure no longer affects the value of capacitance of these dummy taxels that depends only on the temperature, therefore their readings can be used to compensate the thermal drift of the remaining 10 taxels (see Section~\ref{sub:tempcomp}). Since the surface of the FPCB had to host only 10 taxels we could also increase their size (from 12.56 $mm^{2}$ to 15.20 $mm^{2}$, i.e. 20\%). We also rounded the tips of the triangles to increase the signal to noise ratio and simplify the integration process, since triangular modules with smooth edges are easier to glue.

\begin{figure}[!t]
\centering
\includegraphics[width=\columnwidth]{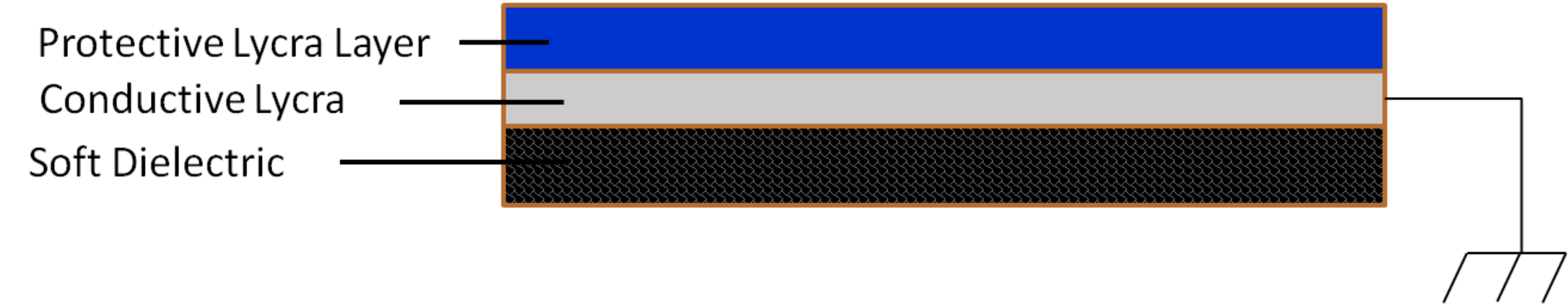}
 \caption{The sandwich of fabric that constitutes the new dielectric for the sensor. The deformable bottom layer is made with 3D air mesh fabric, the medium layer is made with conductive Lycra that acts as common ground plane for all the taxels. The third layer protects the Lycra and improves the mechanical property of the sensor.}
\label{fabric}
\end{figure}

The second major improvement of the tactile system is in the technique used to build the layer covering the FPCB. To improve the robustness of the skin system and simplify its integration process we replaced conductive Lycra and silicone foam with a sandwich made of 3D fabric used for clothing\footnote{3D air mesh fabric}, Lycra and protective fabric (see Figure \ref{fabric}). The advantage is that the 3D fabric is more flexible than the silicone (corresponding to larger sensitivity, see section \ref{sub:sens}) and that the gluing of the different layers and the production process is made using industrial techniques and machines.

\subsection{Integration of the capacitive skin on the iCub}

\begin{figure}[b]
\centering

\subfigure[]{
\includegraphics[width=1.5 in, height = 1.2 in]{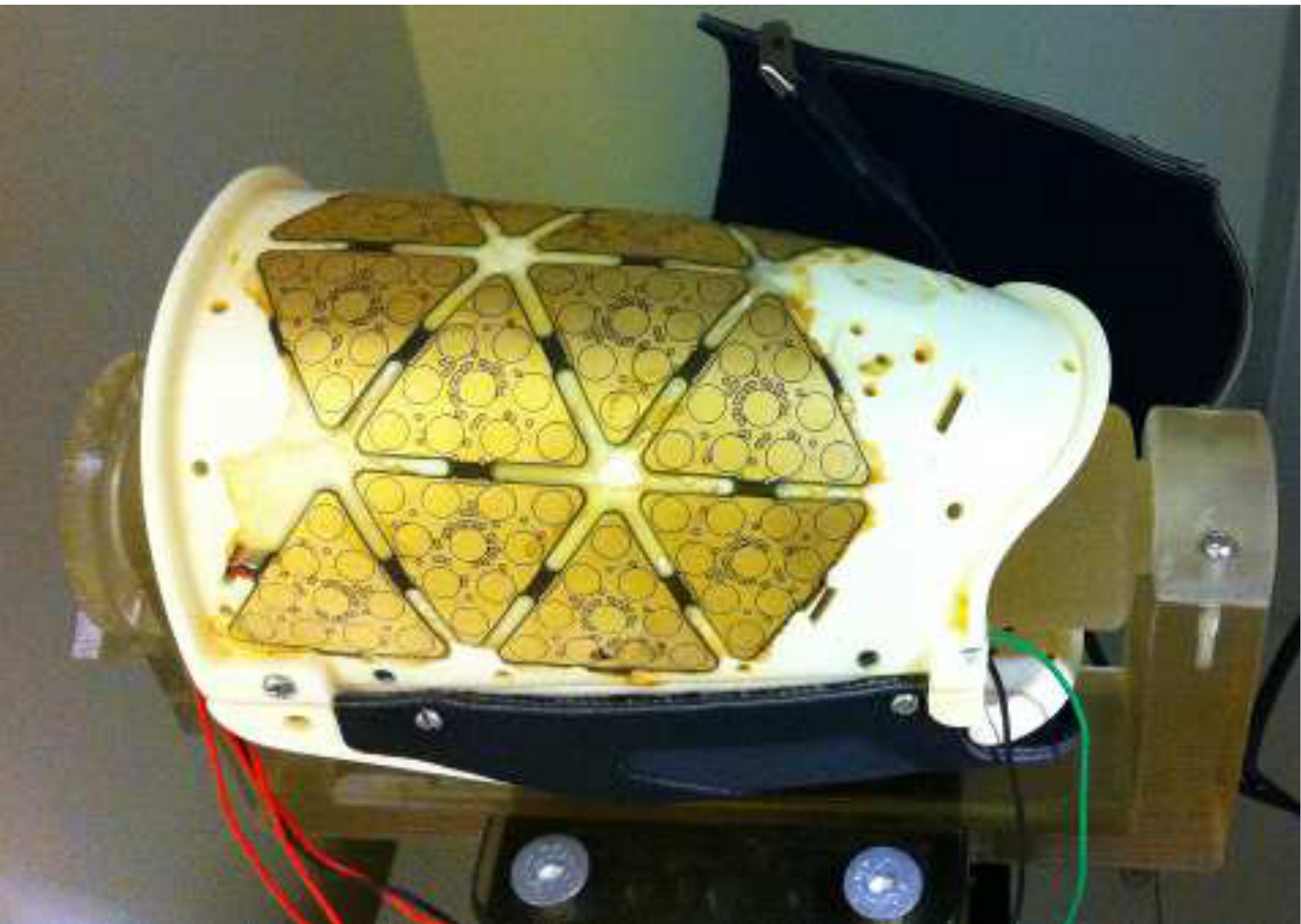}\label{upperarm1}}
\hspace{0.25mm}
\subfigure[]{
\includegraphics[width=1.5 in,height = 1.2 in]{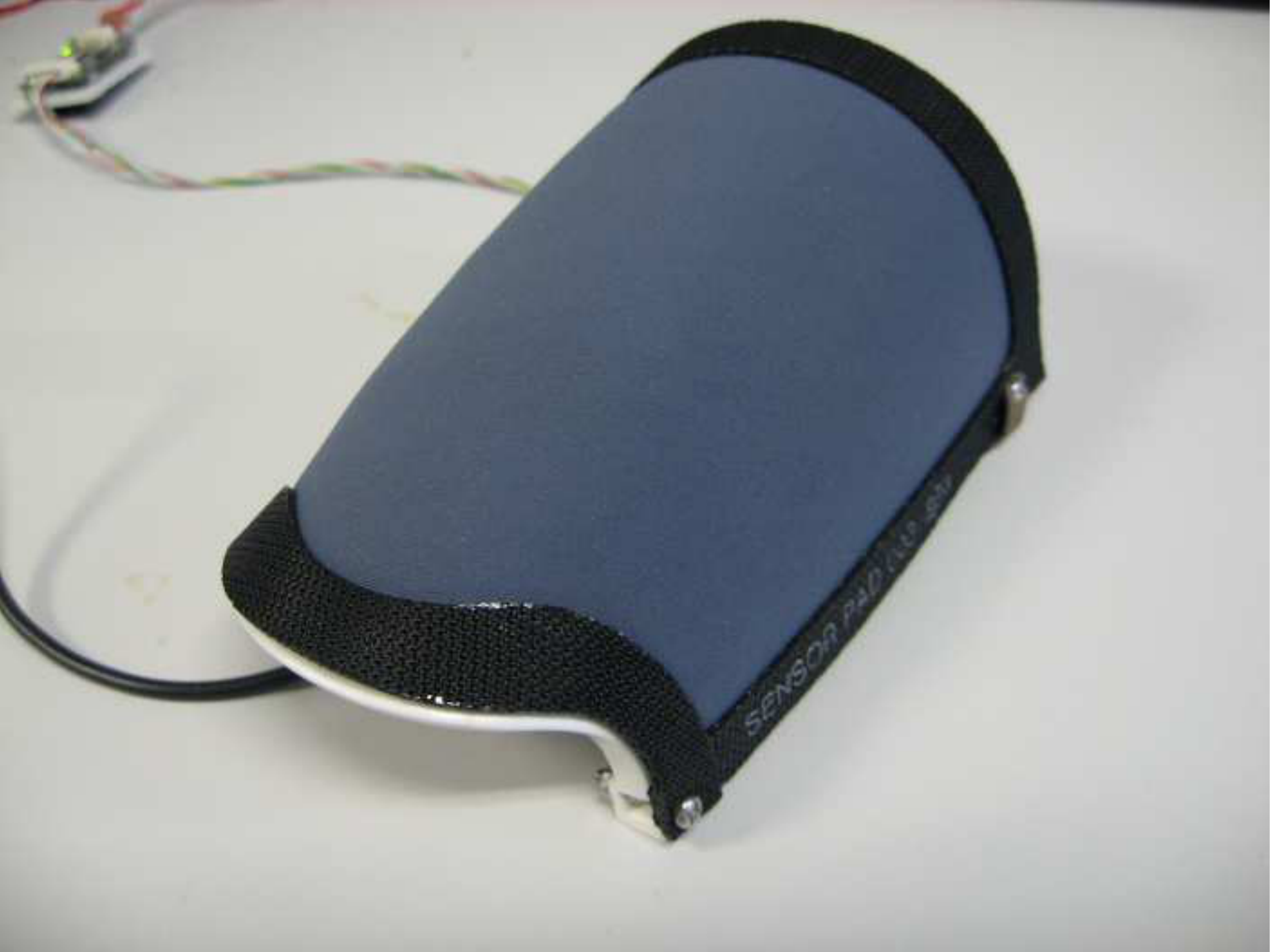}\label{upperarm2}}
\hspace{0.25mm}
\caption{The integration of the sensor on the iCub forearm: (a) the sensor patch is first glued on top of the iCub forearm cover then, (b) the dielectric layer is fixed on top of the sensor with screws.}
\label{newskin}
\end{figure}

The integration procedure was deeply simplified by the use of the new dielectric layer. One of the problems with capacitive sensors is that when they are deployed on a non-flat surface the elastomeric medium bends. This introduces unwanted strain in the material that reduces the dynamical resolution of the sensor. To avoid this, we preform the sensor using a thermoforming procedure to guarantee better adhesion to the part that must be covered. The external layer has special hemlines with holes that host screws to keep the cover in place (see Figure~\ref{upperarm2}). Therefore it can be easily substituted, if damaged, and removed to check the status of the FPCB and electronics below. This is an important improvement since in the previous version the elastomer was strongly glued to the sensor and its removal implied destroying the whole system.

\begin{figure}[t]
\centering

\subfigure[]{
\includegraphics[width=1.5 in]{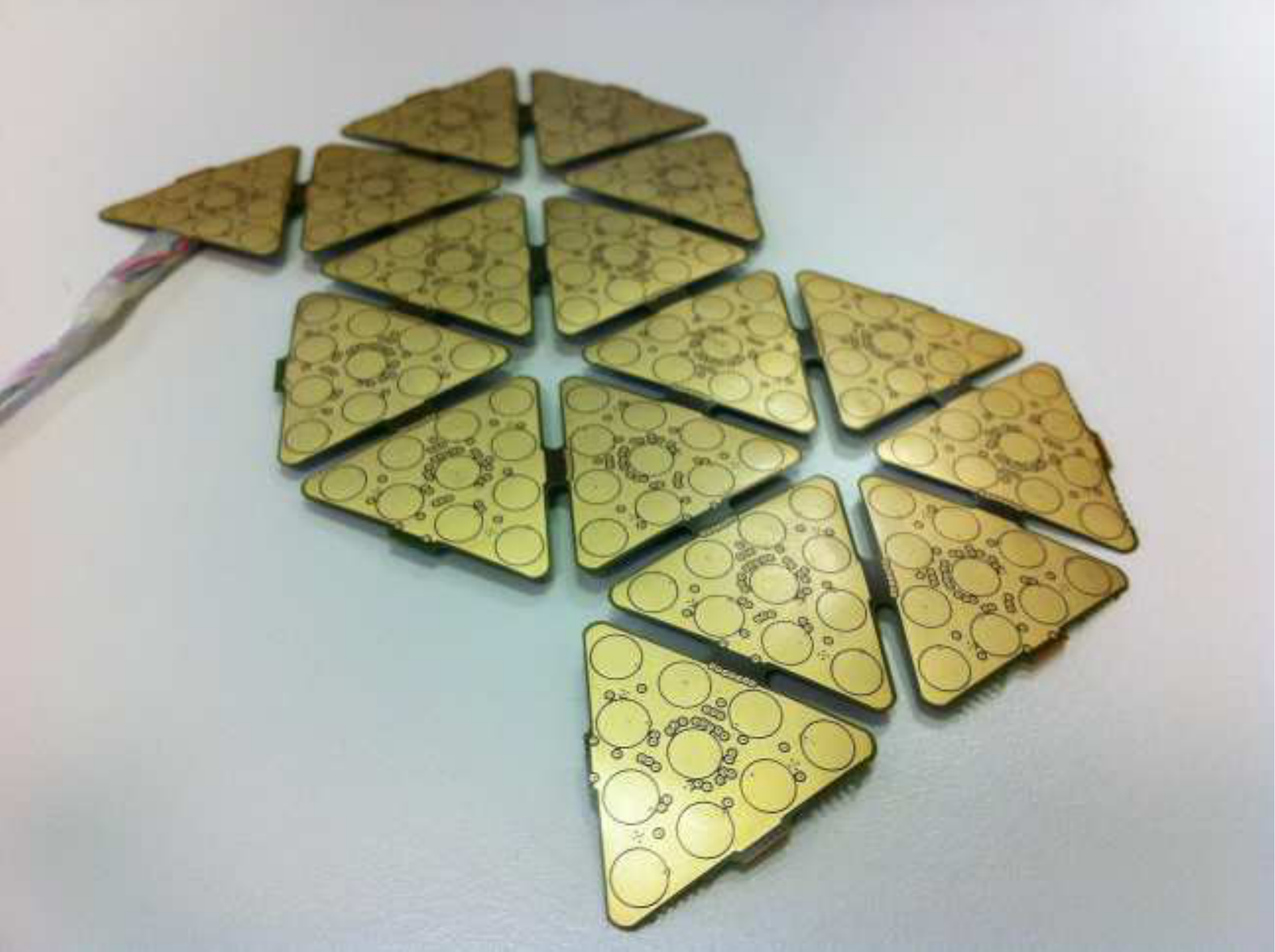}\label{patchfront}}
\hspace{0.25mm}
\subfigure[]{
\includegraphics[width=1.15 in, angle=90]{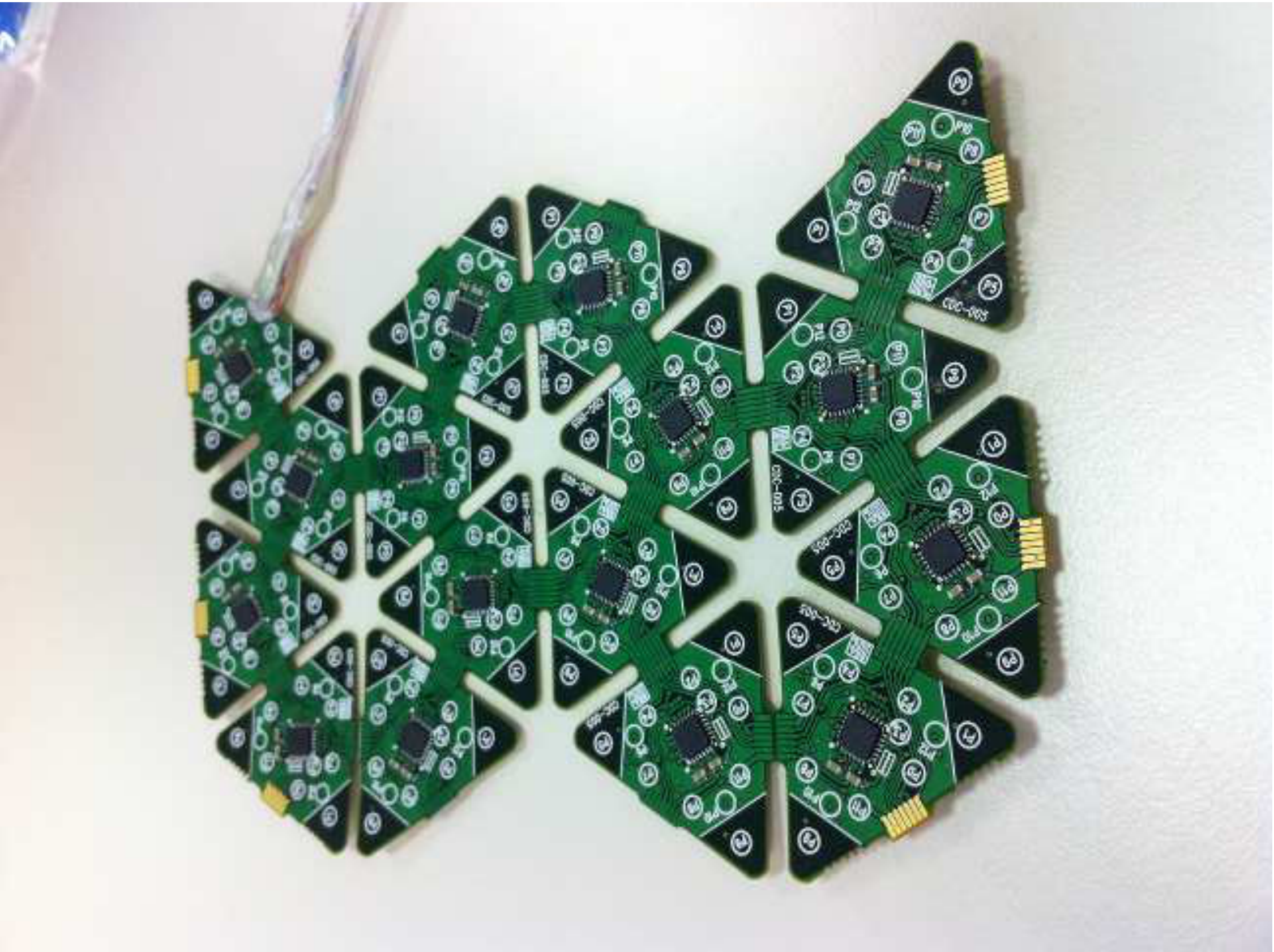}\label{patchback}}\\
\subfigure[]{
\includegraphics[width=1.5 in, height = 1.2 in]{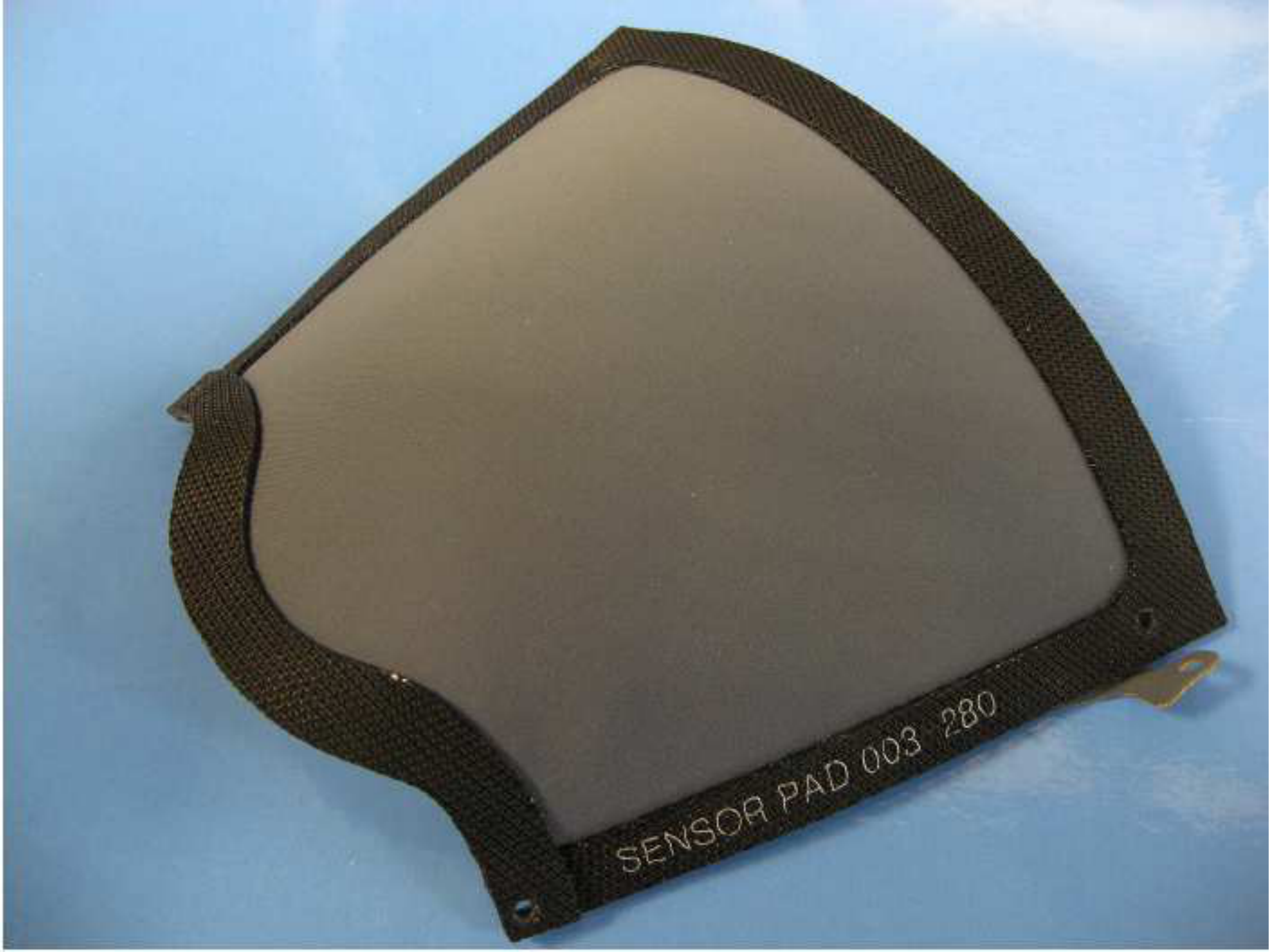}\label{fabricfront}}
\hspace{0.25mm}
\subfigure[]{
\includegraphics[width=1.5 in, height = 1.2 in]{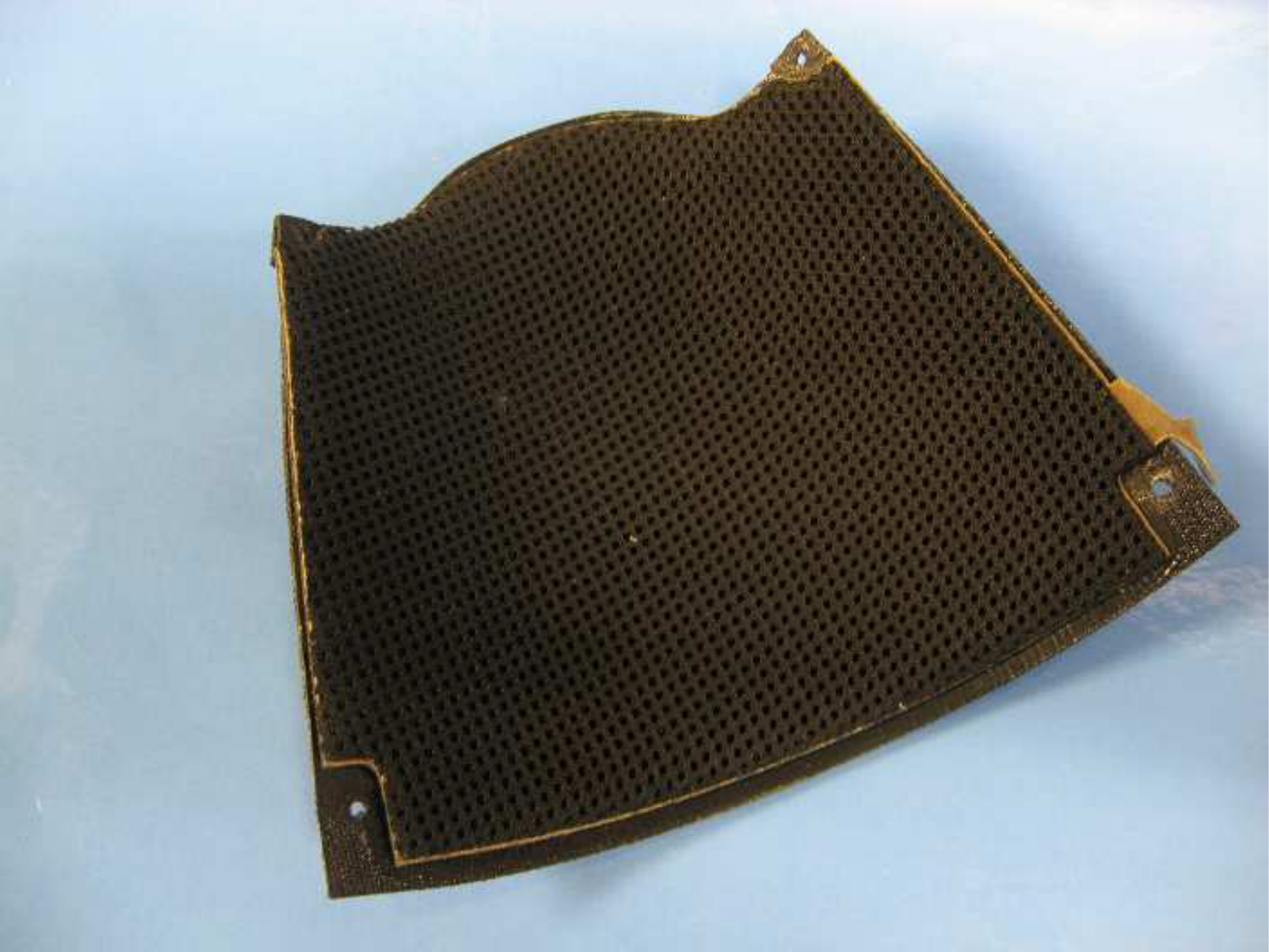}\label{fabricback}}
\hspace{0.25mm}
\caption{The patch and the dielectric layer that have been integrated on iCub forearm. (a) front and (b) back side of the patch (16 triangular modules). (c) the front side of the dielectric cover; it is possible to notice that the cover has been formed to adapt to the shape of the forearm. (d) the back side of the dielectric cover. Here it is possible to notice the grid that characterises the 3D air mesh fabric.}
\label{integration}
\end{figure}

We report here the steps that were followed for the integration of the sensor on the iCub forearm:
\begin{itemize}
\item{The mesh of triangles (see Figure~\ref{patchfront} and \ref{patchback}) was glued on the cover of the iCub forearm using a bi-component glue and with the help of a vacuum system in order to improve the adhesion on the 3D surface. In Figure~\ref{upperarm1} it is possible to see the final result of this procedure.}
\item{For the dielectric layer different layers of fabric have been glued together, cut and shaped to adapt to the robot part (see Figure \ref{fabricfront} and Figure \ref{fabricback}). The cover was then mounted and fixed with screws to the iCub forearm (see Figure~\ref{upperarm2}). As discussed above this allows easy mounting and substitution.}
\end{itemize}

\begin{figure}[t]
\centering

\subfigure[]{
\includegraphics[width=2.5in]{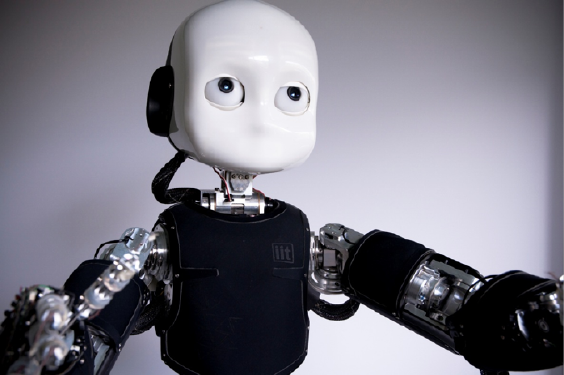}\label{icub}}\\
\subfigure[]{
\includegraphics[width=2.5in]{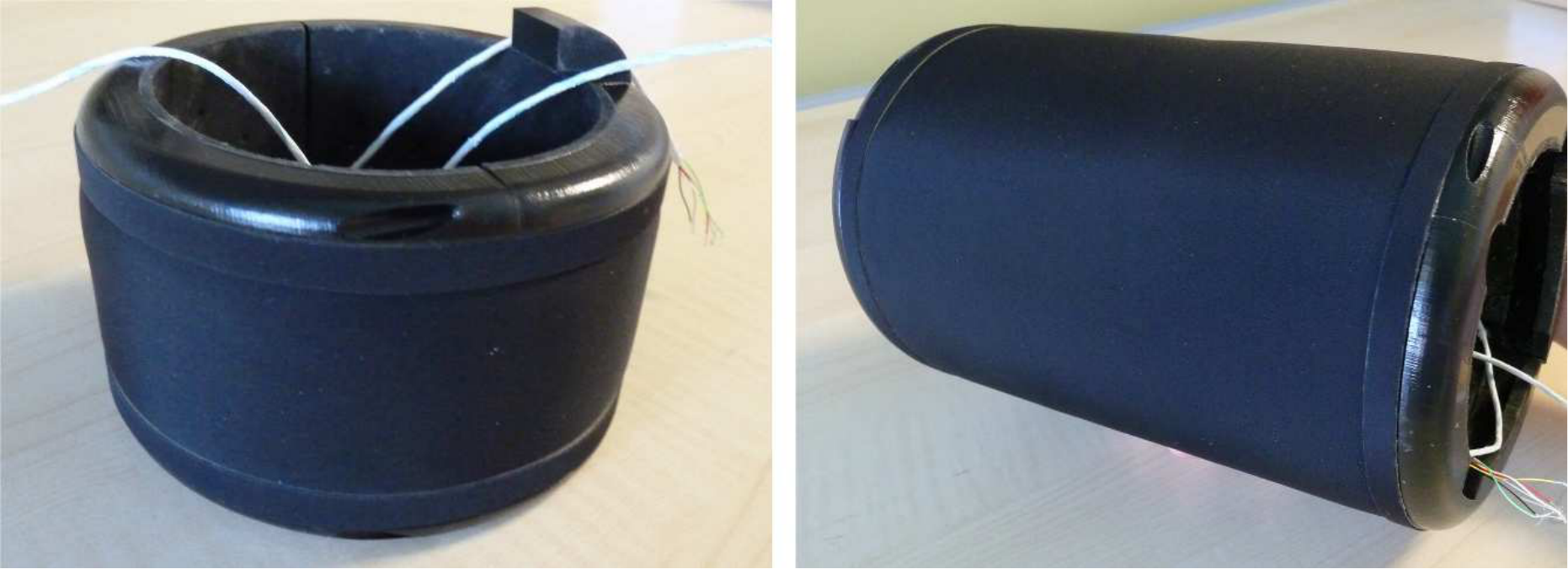}\label{barret}}
\hspace{0.25mm}
\caption{The integration of the sensor on different robots: (a)iCub, (b)WAM arm from Barret Technology}
\label{integration}
\end{figure}

This procedure has been repeated for the integration of the sensor on other iCub parts (the two arms, palms and torso, see Figure~\ref{icub}) and also on the WAM arm from Barret Technology (see Figure~\ref{barret}). Overall the skin system mounted on the iCub has 1868 taxels (104 on the two hands, 610 on the two arms and 440 on the torso), whereas on the WAM arm we mounted 1500 taxels.

\section{EXPERIMENTS}
\label{sec:exp}

\subsection{Experimental Test Setup}

\begin{figure}[t]
\centering
\subfigure[]{
\includegraphics[width=1.2 in, height=2in,]{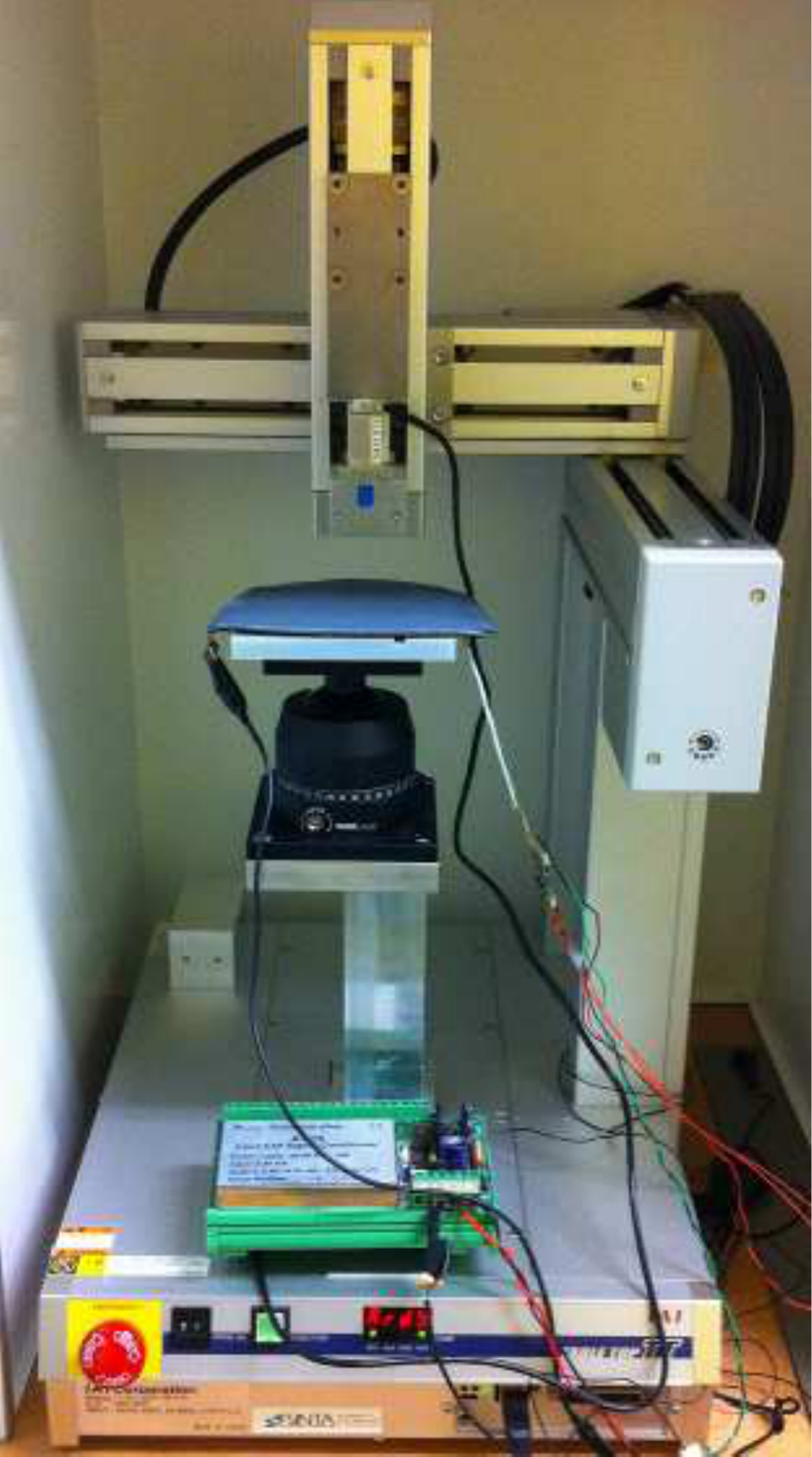}\label{setup}}
\hspace{0.25mm}
\subfigure[]{
\includegraphics[width=1.5 in]{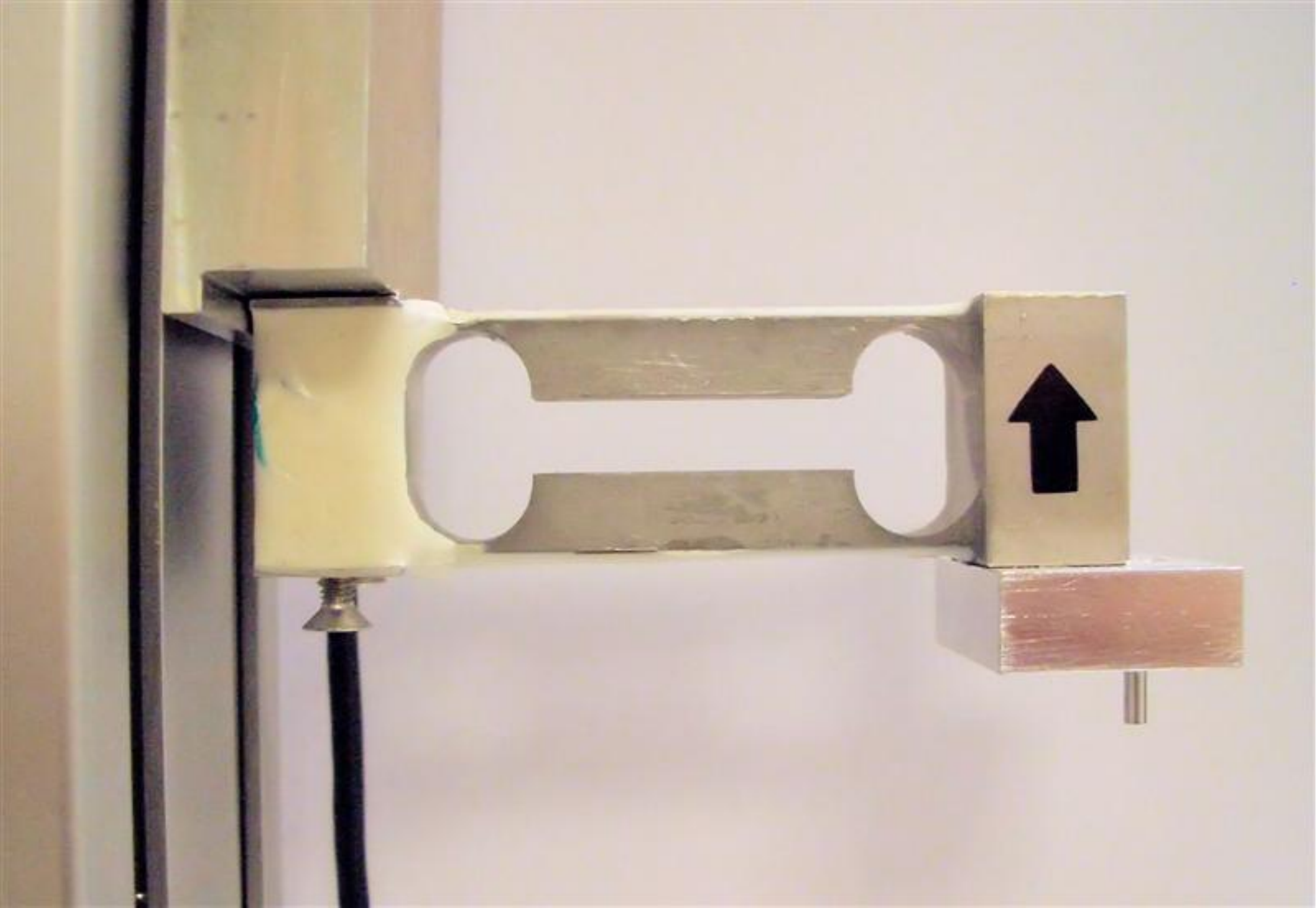}\label{cellagrande}}
\caption{The experimental setup used to characterise the sensor.
The Cartesian robot (TT-C3-2020 from IAI) is shown in (a). In (b) a close-up of the probe and the off-centre loadcell (AS kg 0.5, from
Laumas) is shown.}
\label{expsetup}
\end{figure}

The experimental tests were performed using a 3 axis Cartesian robot (TT-C3-2020 from IAI Inc.). The robot supports an off-center load cell (AS kg 0.5 from Laumas Elettronica S.r.l.) to which cylindrical probes of varying diameters, depending on the experiment, were attached. The robot moved the probe in x, y plane and pushed it vertically against the tactile sensor at different locations.
All the measurements (robot position and load cell values) were collected at the frequency of 25 Hz.
The signal from the load cell was amplified by an AT10 from Precise Instruments Corp. and acquired using the same microcontroller board that was used to send the measurements of the capacitive tactile sensor system to the PC. In this way, we collected synchronised data from the capacitive pressure sensor system and the load cell. The applied pressure was computed as the force measured by the load cell divided by the contact area. Since we measured the capacitance of the sensors we converted the output to Farad using the nominal resolution of the CDC.
Figure \ref{expsetup} shows the measurement setup: Figure~\ref{setup} shows the Cartesian robot whereas Figure \ref{cellagrande} shows the off-center load cell.

\begin{figure}[t]
\centering
\subfigure[]{
\includegraphics[width= 1.5 in]{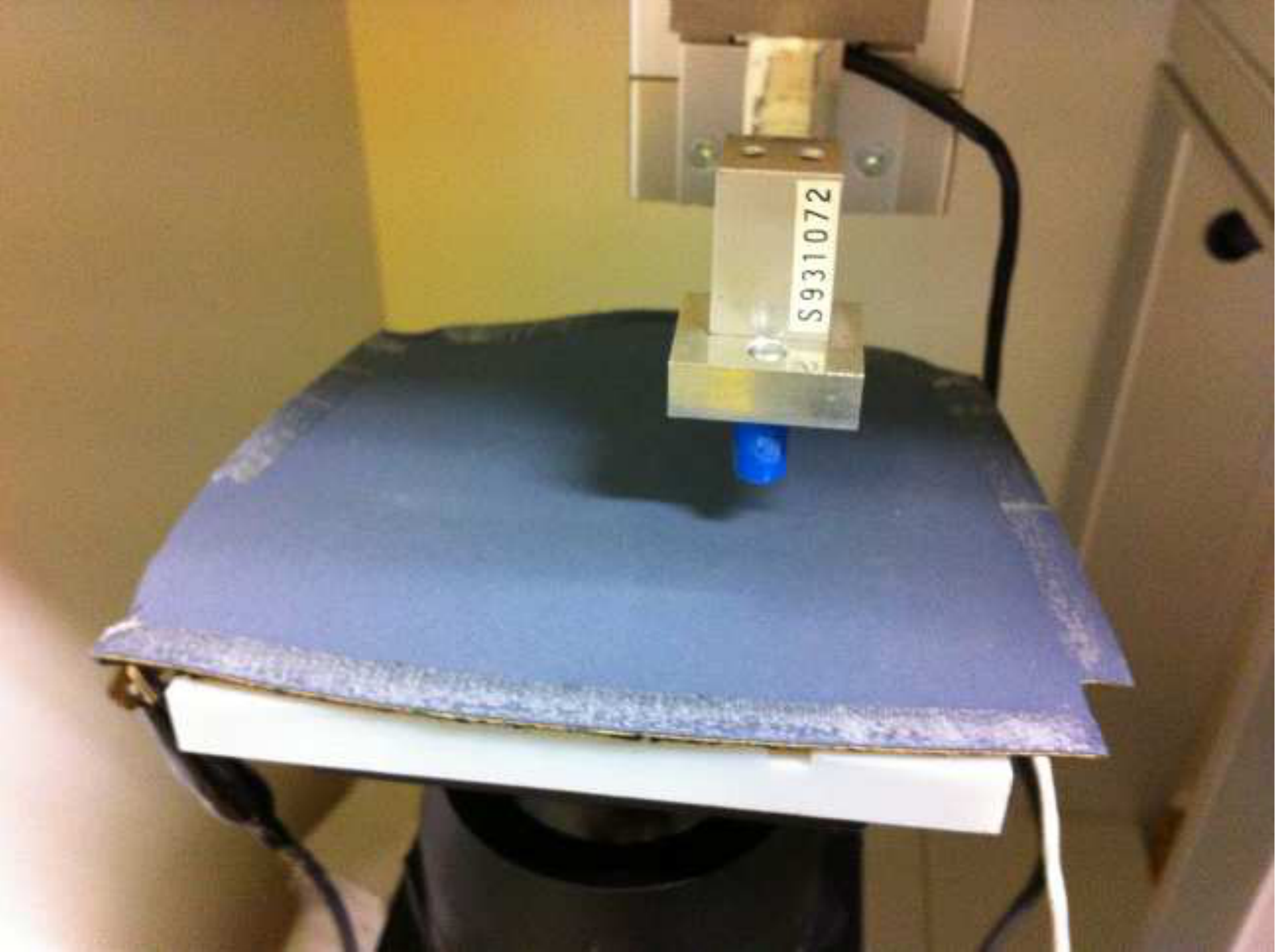}\label{cella}}
\hspace{0.25mm}
\subfigure[]{
\includegraphics[width=1.5 in]{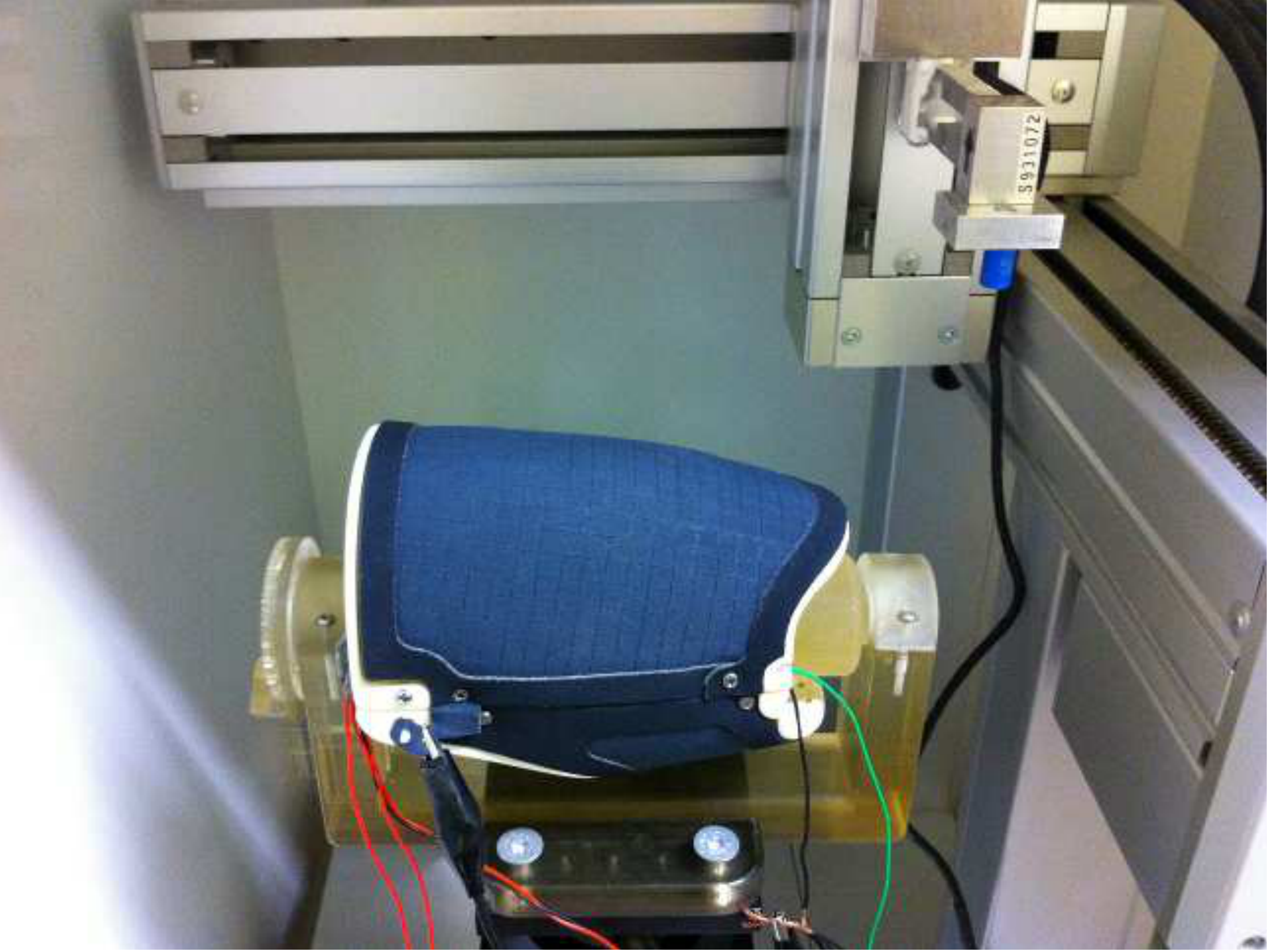}\label{upperarmsetup}}
\caption{The two prototypes used in the experiments. (a) the flat prototype (b) the 3D prototype (the cover of the iCub forearm).}
\label{prototype}
\end{figure}

The experiments were performed on two different prototypes (see Figure~\ref{prototype}). The first prototype consisted in a patch of 16 triangular modules, glued on a flat structure and covered with a dielectric layer with a thickness of 2 mm that, for this particular configuration, was  glued on top of the sensor. The second prototype was a patch of sensors mounted on the iCub forearm (we refer to it as the 3D prototype). Similarly to the flat prototype this part had 16 triangular modules. 
We evaluated the performance of the sensor in terms of sensitivity, repeatability, hysteresis and spatial resolution using the flat prototype, while for the thermal drift we  used the 3D prototype.

\subsection{Sensitivity and Repeatability}
\label{sub:sens}

\begin{figure}[!t]
\centering
\includegraphics[width=2.5in]{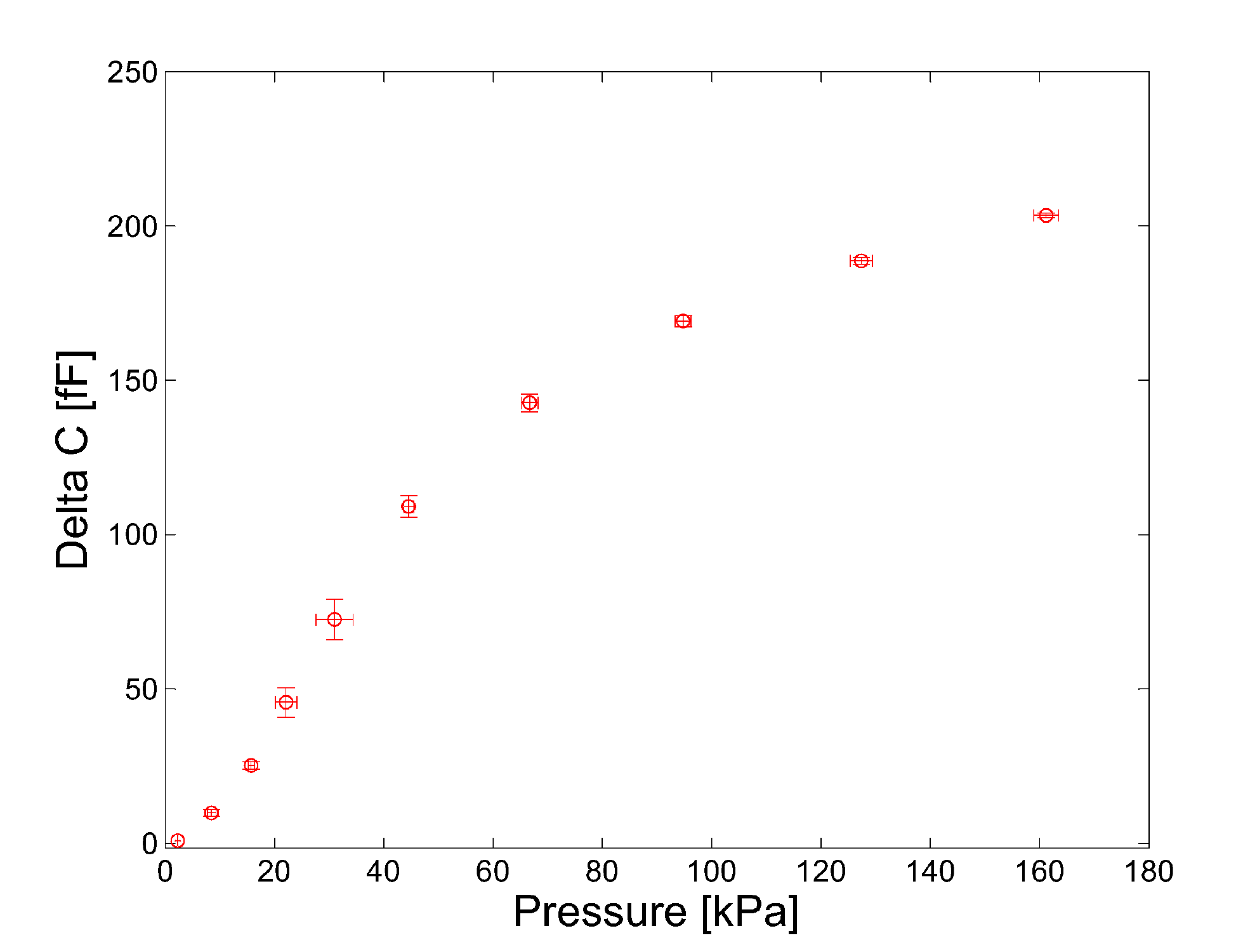}
 \caption{This plot shows the average response of one of the excited taxel, with a probe of 7 mm diameter, for different pressure (circles) with standard deviations (bars) across 15 cycles.}
\label{repeatability}
\end{figure}

In order to evaluate sensitivity and repeatability of the sensor, we collected data from the central taxel of two different triangular modules. We performed experiments with probes of different diameters (i.e. 2 mm and 7 mm). The experiments were conducted in the following way: The probe moved down to a specified z-position as fast as possible, remained there for two seconds and it moved up again to the initial no-contact position. Subsequently it moved down again, this time 0.2 mm deeper than before. The whole process was repeated until the probe had pushed to the deepest defined point corresponding to the maximum force detectable by the load cell. We conducted this experiment 15 times for each taxel. Between each repetition we waited 15 minutes to remove any hysteresis effect. 
We report in Figure~\ref{repeatability} the response of one of the excited taxel and the standard deviation during all the experiments. This result shows that the response of the sensor is repeatable.

\begin{figure}[!t]
\centering
\includegraphics[width=2.5in]{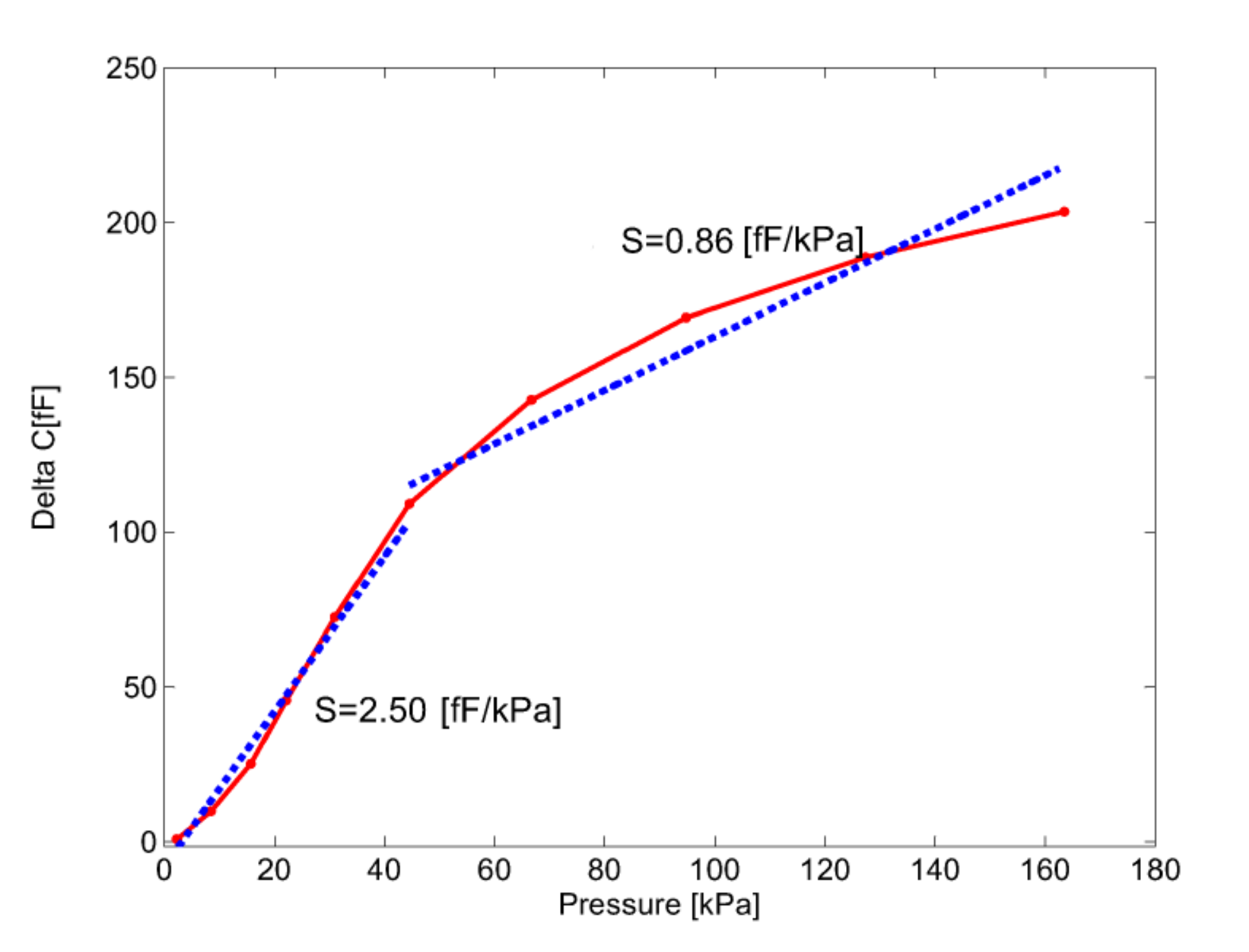}
 \caption{The sensitivity of the sensor. Continuous curve: interpolation of the average of the sensor response for all the experiments (the average of the experimental values are indicated by markers). Two different linear ranges of pressures have been identified for the calculation of the sensitivity. Dotted line: the sensitivity for the two identified ranges (see text for details).}
\label{sensitivity}
\end{figure}

\begin{figure}[!t]
\centering
\includegraphics[width=2.5in]{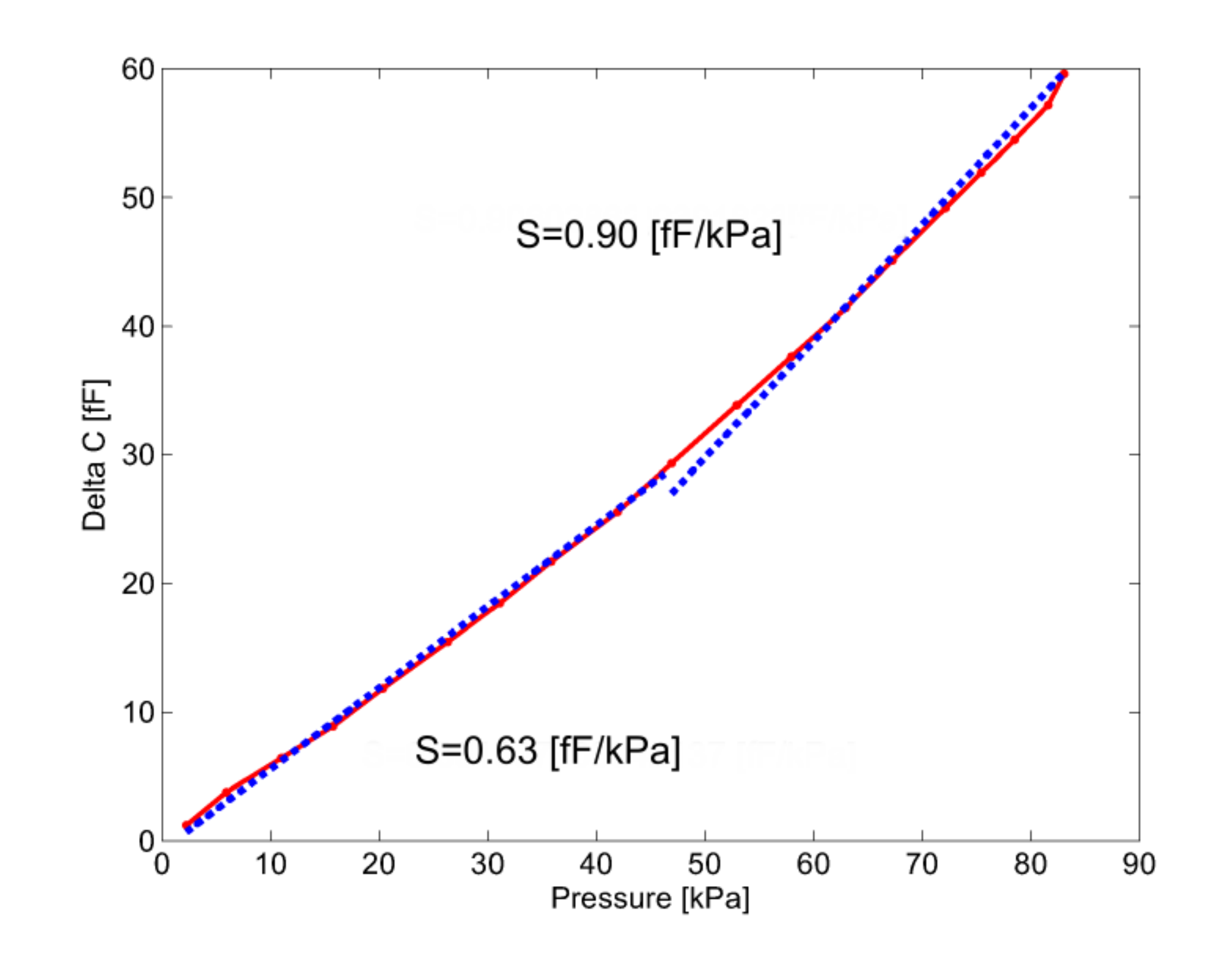}
 \caption{The sensitivity of the previous version of the sensor that used silicone foam as dielectric layer~\cite{Schmitz08}. Continuous curve: interpolation of the average of the response for all the experiments. Dotted line: the sensitivity for the two ranges. Notice that for the same range of pressures the sensitivity in the new sensor is higher (compare with Figure~\ref{sensitivity}).}
\label{sensitivitysh}
\end{figure}

We define the average pressure sensitivity S as $\frac{\Delta C}{\Delta P}$, where $\Delta C$ denotes the variation of capacitance and $\Delta P$ the variation of the applied pressure. Figure~\ref{sensitivity} reports the response of the sensor and the pressure sensitivity S (the dotted line) over two adjacent pressure ranges. The first linear region included pressures from 2 to 45 [kPa], whereas the second region included the range from  65 to 160 [kPa]. In these linear areas sensitivity was estimated to be respectively $S=2.50~[fF/kPa]$ and $S= 0.86~[fF/kPa]$.
As it is possible to notice the sensitivity is higher in the first linear range and after it decreases as the dielectric layer gets compressed and becomes less compliant. 
For comparison, in Figure~\ref{sensitivitysh}, we report the sensitivity of the previous version of the skin system~\cite{Schmitz08} that employed a soft foam elastomer as dielectric (SomaFoama from Smooth-on\footnote{www.smooth-on.com}). Despite the low Young's modulus of the foam elastomer (i.e. 150~[kPa]), the sensitivity is lower in the pressure range 2-45~[kPa] ($S= 0.63 [fF/kPa]$). This shows that the new dielectric layer leads to a higher sensitivity for the sensor, especially in the preferable lower range of pressures.

\subsection{Hysteresis and Relaxation}

\begin{figure}[!t]
\centering
\includegraphics[width=2.5in]{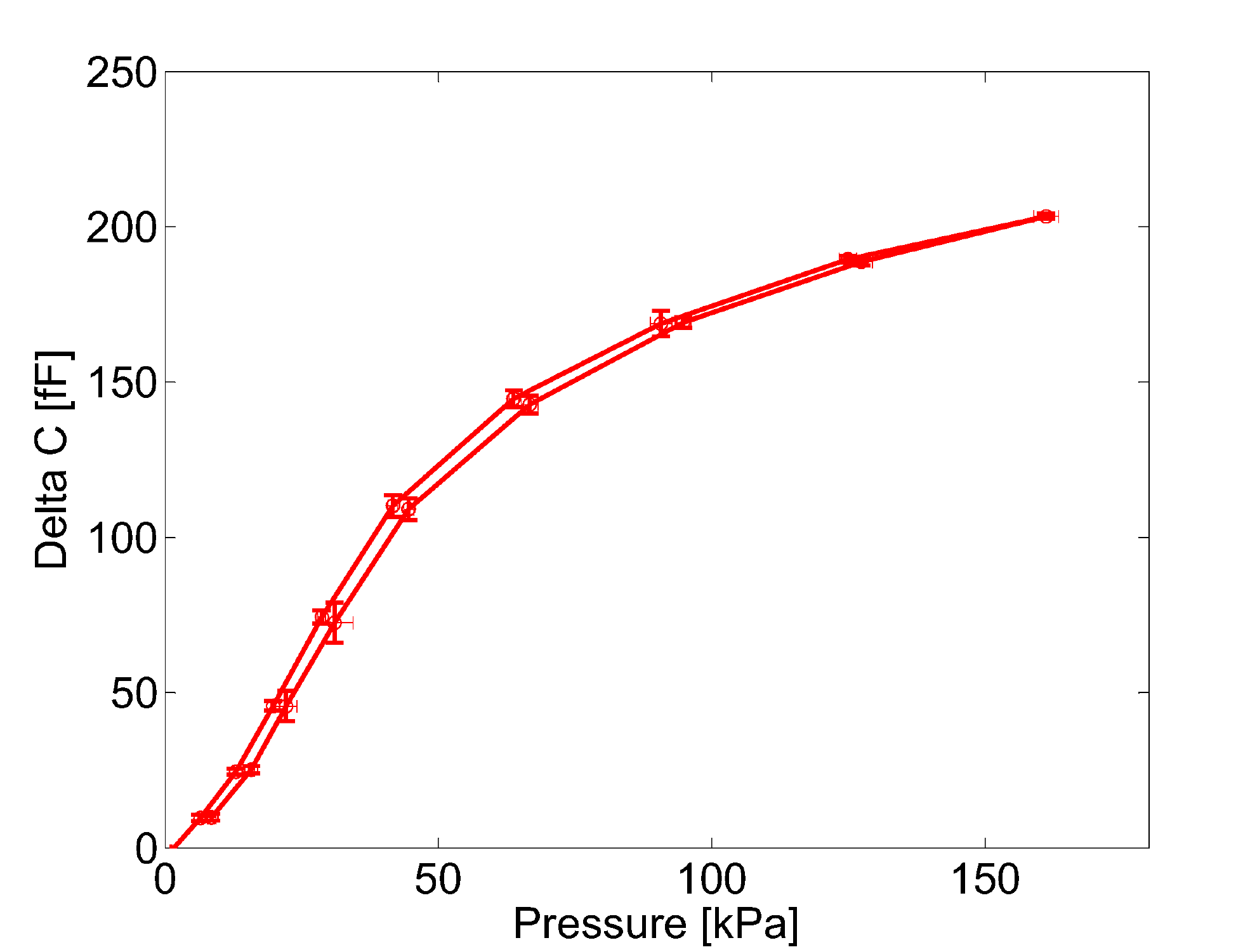}
 \caption{Response of one of the excited taxel (the same as for the repeatability experiment) acquired in the hysteresis experiment; the average and the standard deviation (bars) are shown.}
 \label{hysteresis}
\end{figure}

In order to evaluate the hysteresis of the sensor, we excited the same taxels as in the previous test with the same load-unload pressure cycle. However, in this case we performed the experiments 15 times for each taxel waiting only 1 minute between each repetition. 
Figure \ref{hysteresis} shows for each step the average measurement of one of the excited taxel, together with the standard deviation. There is indeed a certain amount of hysteresis since the responses do not overlap exactly.
However the hysteresis is quite low: the maximum difference between cycles corresponds to a pressure of 28.6 kPa and is about 9.1 fF (corresponding to roughly $5\%$ of the whole range). This result is remarkable, especially for a capacitive sensor.

\begin{figure}[!t]
\centering
\includegraphics[width=2.5in]{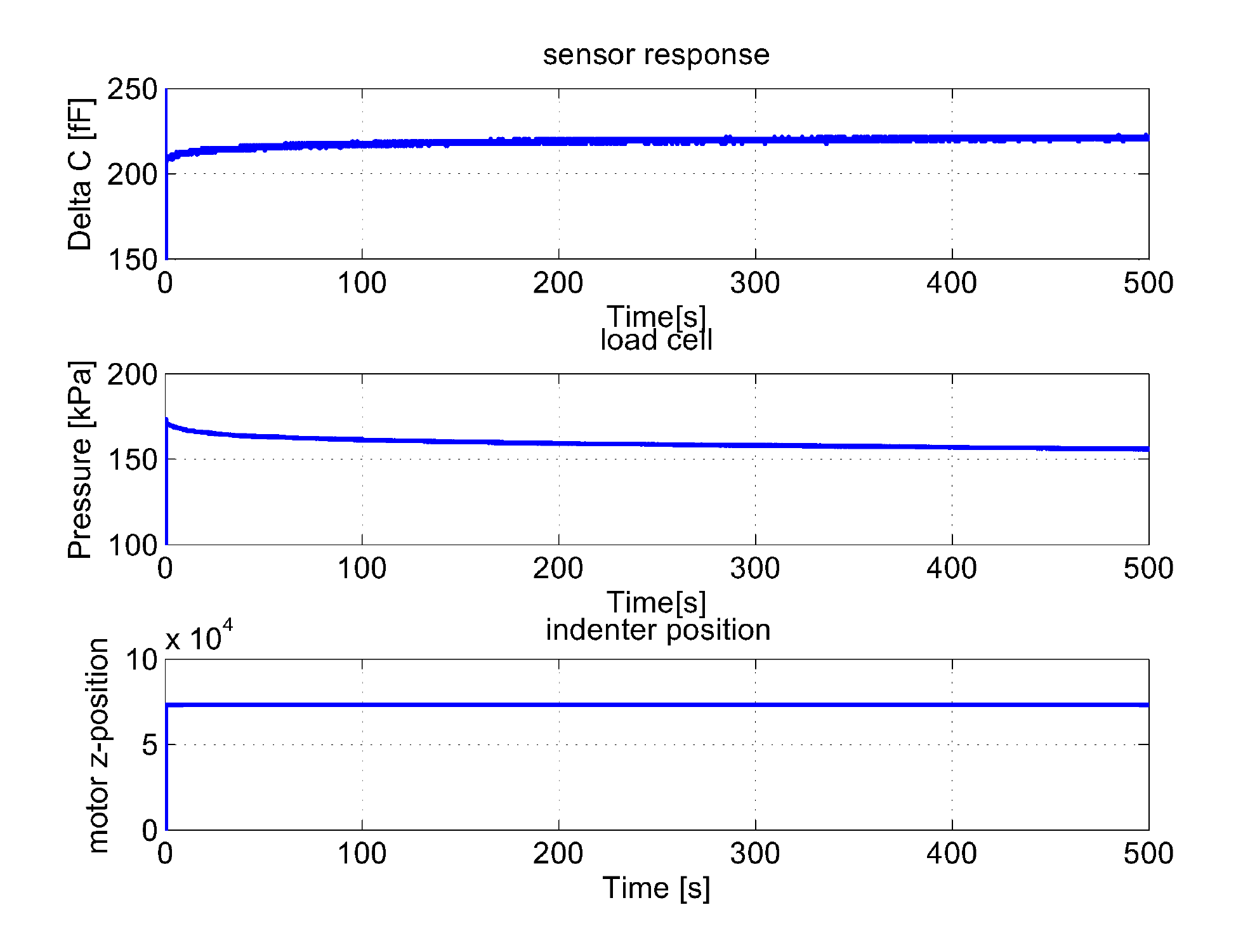}
 \caption{Stress relaxation test. From top to bottom: sensor response $\Delta C$, load cell response and indenter (7 mm diameter) position.}
\label{relax}
\end{figure}

The evaluation of the relaxation of the dielectric layer was performed by forcing a fixed deformation to the dielectric layer for 10 minutes at a constant room temperature.
Figure~\ref{relax} reports the sensor response ($\Delta C$), the load cell response and the position of the indenter. The response of the sensor and the load cell show that there is indeed a stress relaxation (corresponding to an increase of strain) due to the rearrangement of the polymeric chain that leads the material to adapt to the imposed deformation. 
Since the relaxation has an exponential behaviour, we characterised the relaxation constant of the material through the following equation \cite{nielsen}:
\begin{equation}
\label{eq:relax}
\sigma(t)=\sigma_{0}\exp\left(\frac{-t}{\tau}\right)
\end{equation} 
where $\tau$ is the relaxation constant, $\sigma_{0}$ is the initial stress and $\sigma(t)$ is the stress calculated in $t$. In our case we determined $\tau$ equal to 1h and 18min. This parameter can be used to estimate the stress relaxation after a certain time $T$ following constant deformation of the dielectric (i.e. a contact event for an extended amount of time).

\subsection{Spatial Resolution}

\begin{figure}[!t]
\centering
\includegraphics[width=2.5in]{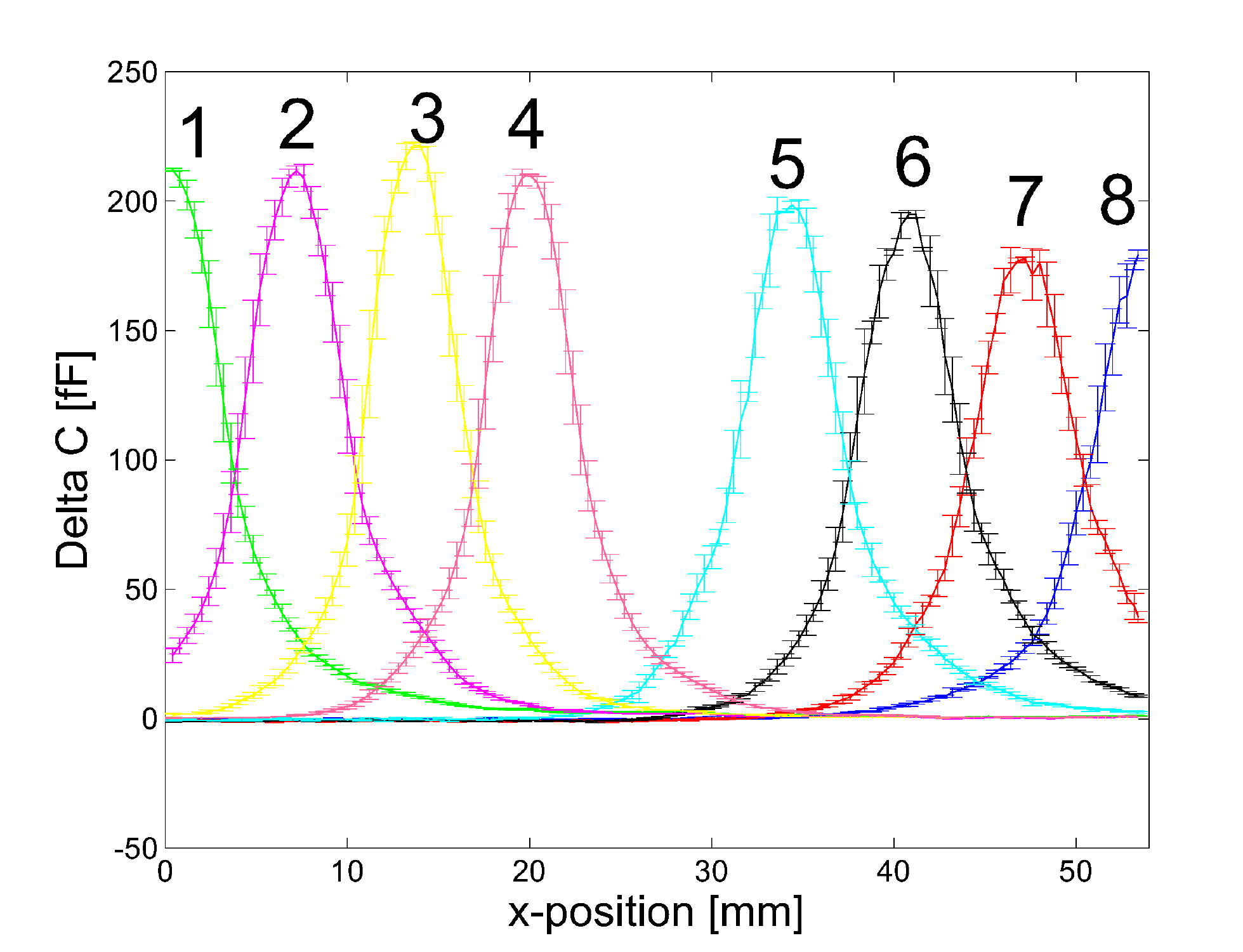}
 \caption{Spatial resolution of the tactile system. The probe (2 mm diameter) pushed the patch of sensors at different positions (each at the relative distance of 0.2 mm), along
a straight line. We show average and standard deviation of the response of all taxels that were activated in the experiment. The numbers represent the different taxels as indicated in Figure~\ref{pspaziale}}
\label{indenter2}
\end{figure}

\begin{figure}[!t]
\centering
\includegraphics[width=2.5in]{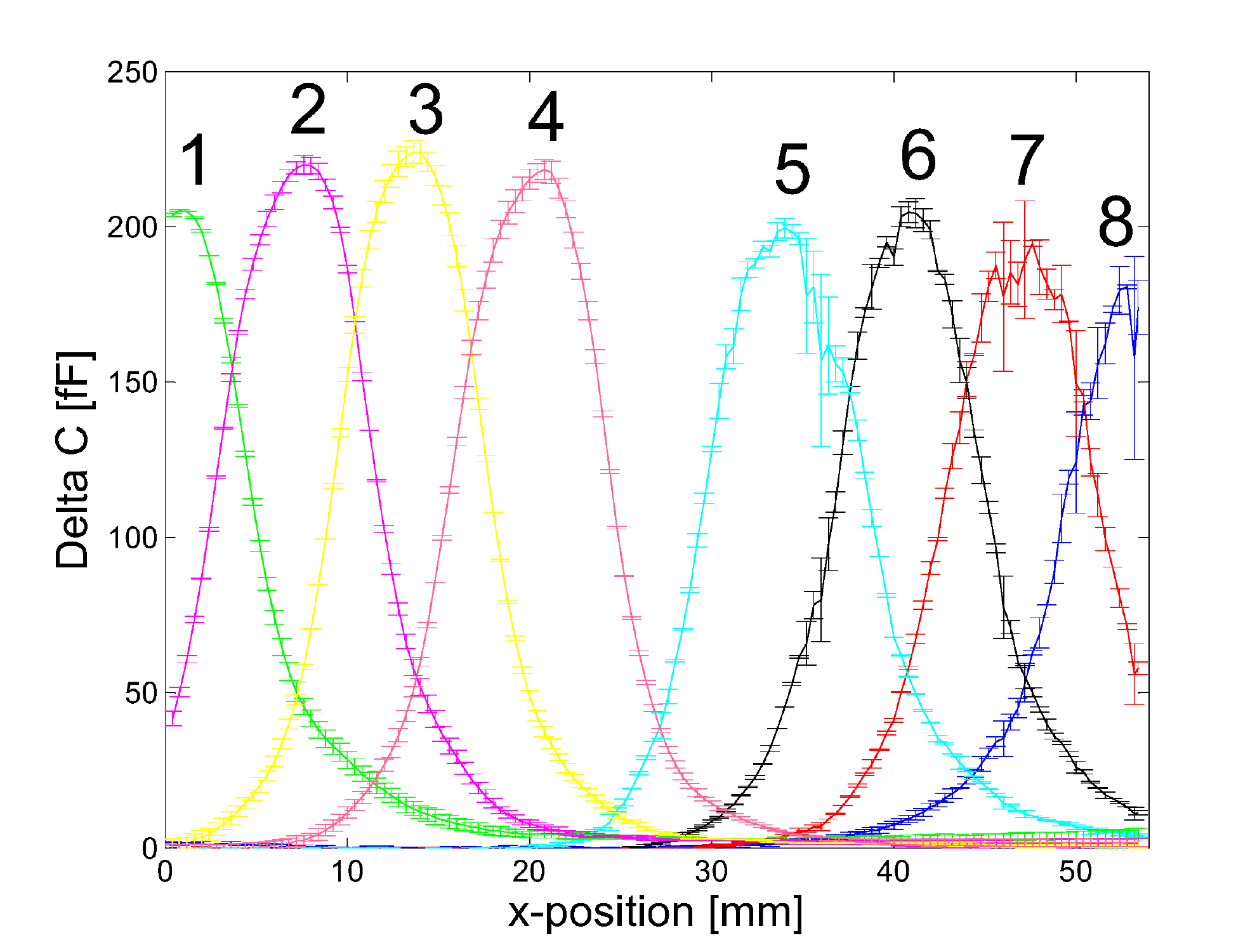}
\caption{Spatial resolution of the tactile system. The probe (7 mm diameter) pushed the patch of sensors at different positions (each at the relative distance of 0.2 mm), along a straight line. We show average and standard deviation of the response of all taxels that were activated in the experiment. The numbers represent the different taxels as indicated in Figure~\ref{pspaziale}}
\label{indenter7}
\end{figure}

\begin{figure}[!t]
\centering
\subfigure[]{
\includegraphics[width=1.7in]{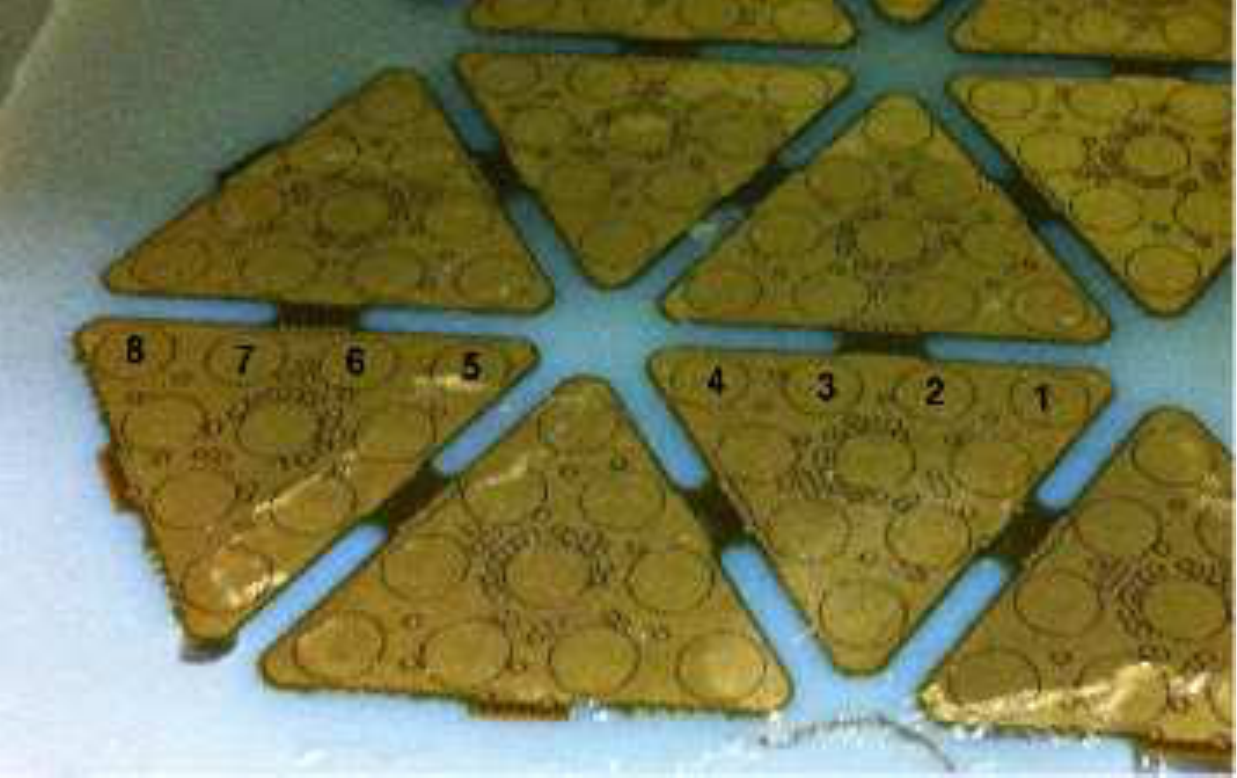}\label{pspaziale}}
\hspace{0.25mm}
\subfigure[]{
\includegraphics[width=1.3in]{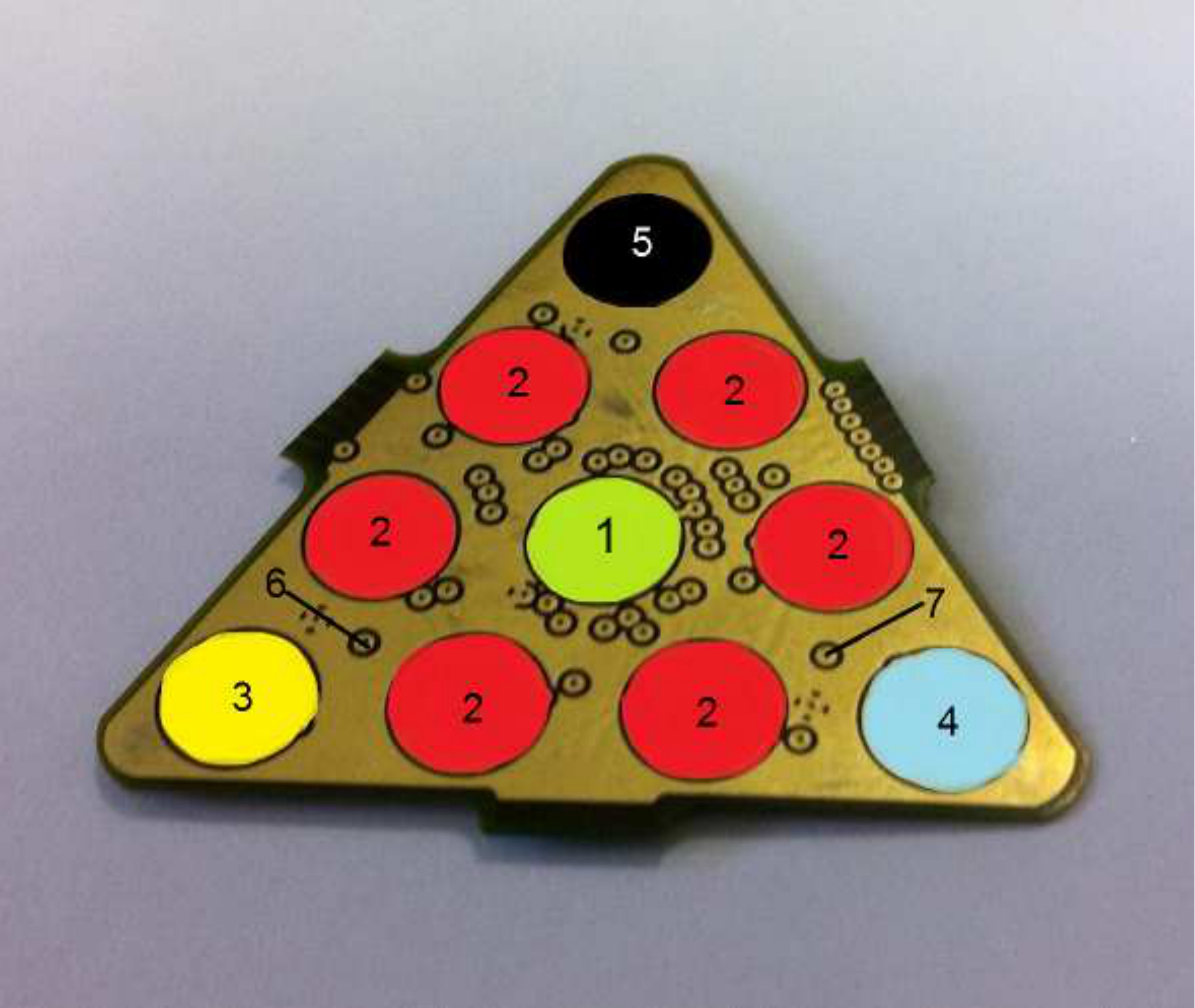}\label{triangolo}}
 \caption{a) The Figure reports the taxels that were excited during the spatial resolution experiment. The numbers identify the order in which the taxels have been excited. b) Representation of the 10 taxels in the triangle, number code. Taxels with similar temperature drift have been grouped together and represented with the same number. The thermal pads are identified by number 6 and number 7.}
\label{taxelsex}
\end{figure}

The spatial resolution of the sensor was evaluated by performing the following experiment. The indenter applied the same pressure to the sensor at different positions along a straight line (corresponding to the the x-axis of the Cartesian robot). We collected measurements in steps of 0.2 mm, but the pressure was not applied consequently at two adjacent positions to avoid the influence of hysteresis on the measurements. To this aim the indenter moved first in steps of 0.4 mm from the initial to the final position and then it moved back from the final to the initial position in steps of 0.4 mm but with an offset of 0.2 mm. At each position, the indenter moved down to a defined z-position and then moved up and subsequently it changed the position. The experiments were repeated three times to confirm the repeatability of the sensor response and with indenters of different diameter (i.e. 2 mm and 7 mm). In Figure~\ref{indenter2} and Figure \ref{indenter7} we show the response of all the excited taxels (Figure~\ref{pspaziale} shows which taxels were excited and in which order) for the indenter with 2 mm of diameter and 7 mm of diameter. The results show that the taxels respond with a bell shaped curve and that the receptive fields overlap. Furthermore it is possible to notice that the width of the bell shaped curves change according to the diameter of the indenter confirming that the deformation of the dielectric layer distributes the pressure to nearby taxels and influences the spatial resolution of the sensor.  Interestingly, recent work demonstrated that this type of response is beneficial since it allows hyperacuity~\cite{Lepora12}.

\subsection{Temperature Compensation}
\label{sub:tempcomp}
As discussed in the introduction one of the improvements of the new version of the tactile system was that the layout of the flexible FPCB was changed to embed two \emph{thermal} sensors to be used for temperature compensation. 
Temperature compensation can therefore be achieved by adding to the taxels output the $\Delta C$ value of one of these \emph{thermal} sensors (the two sensors have almost the same behavior) multiplied by a gain, as reported in the following equation:

\begin{equation}
\label{eq:compensation}
\hat{T}_{i}(n)= T_{i}(n)- K_{i}[T_{h}(n)-\bar{T}_{h}]
\end{equation}
   
Where $\hat{T}_{i}(n)$  is the n-th sample of the value of the taxel \textit{i} after compensation, $T_{i}(n)$ is the raw value of the same taxel, $T_{h}(n)$ 
is the n-th sample of the thermal sensor, $\bar{T}_{h}$ is the average of the values of the thermal sensor at start-up (baseline) and $K_{i}$ is an appropriate gain factor to be estimated during a calibration procedure for each taxel \textit{i}.

\begin{figure}[t]
\centering
\subfigure[]{
\includegraphics[width=2.5in]{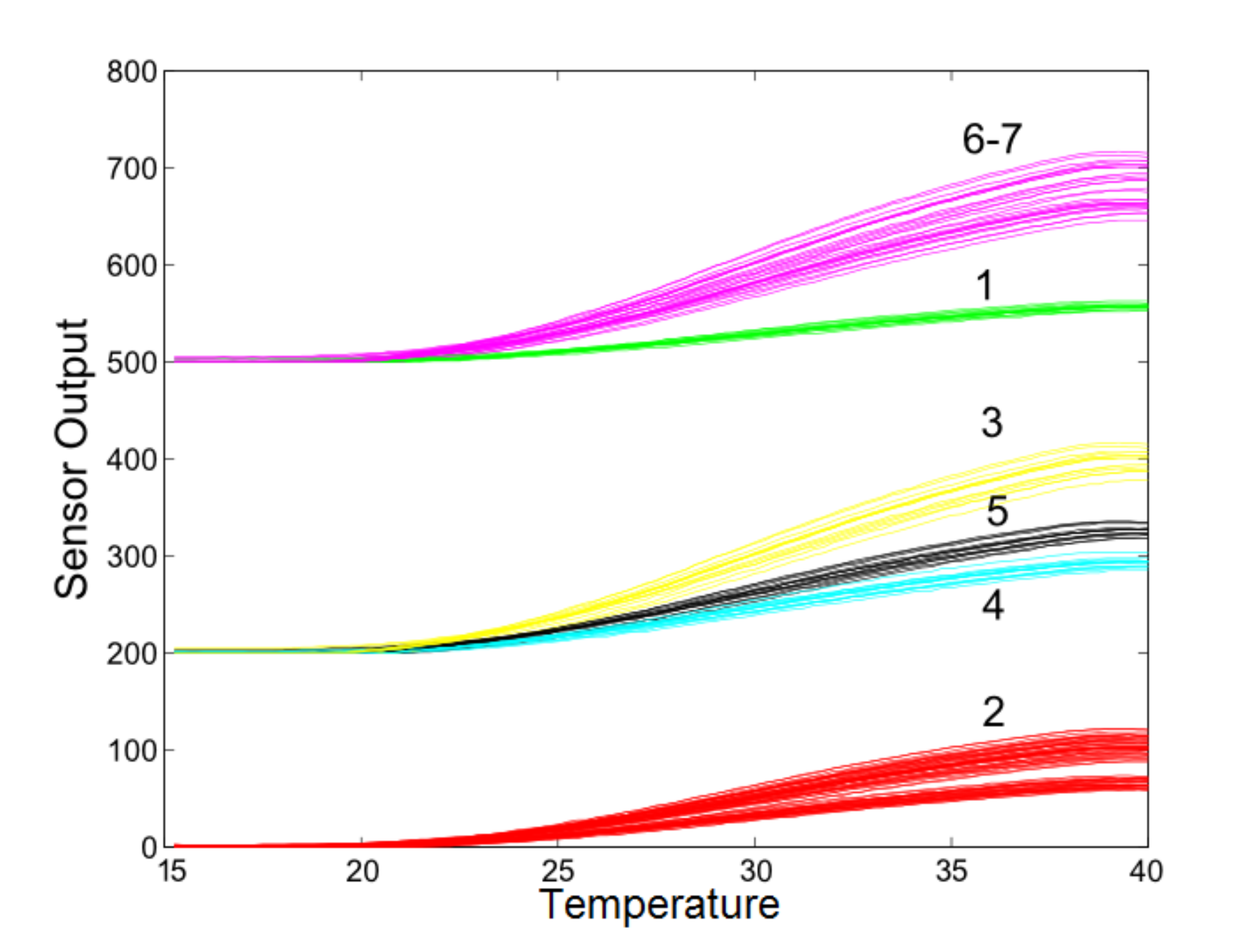}\label{1540}}
\hspace{0.25mm}
\subfigure[]{
\includegraphics[width=2.5in]{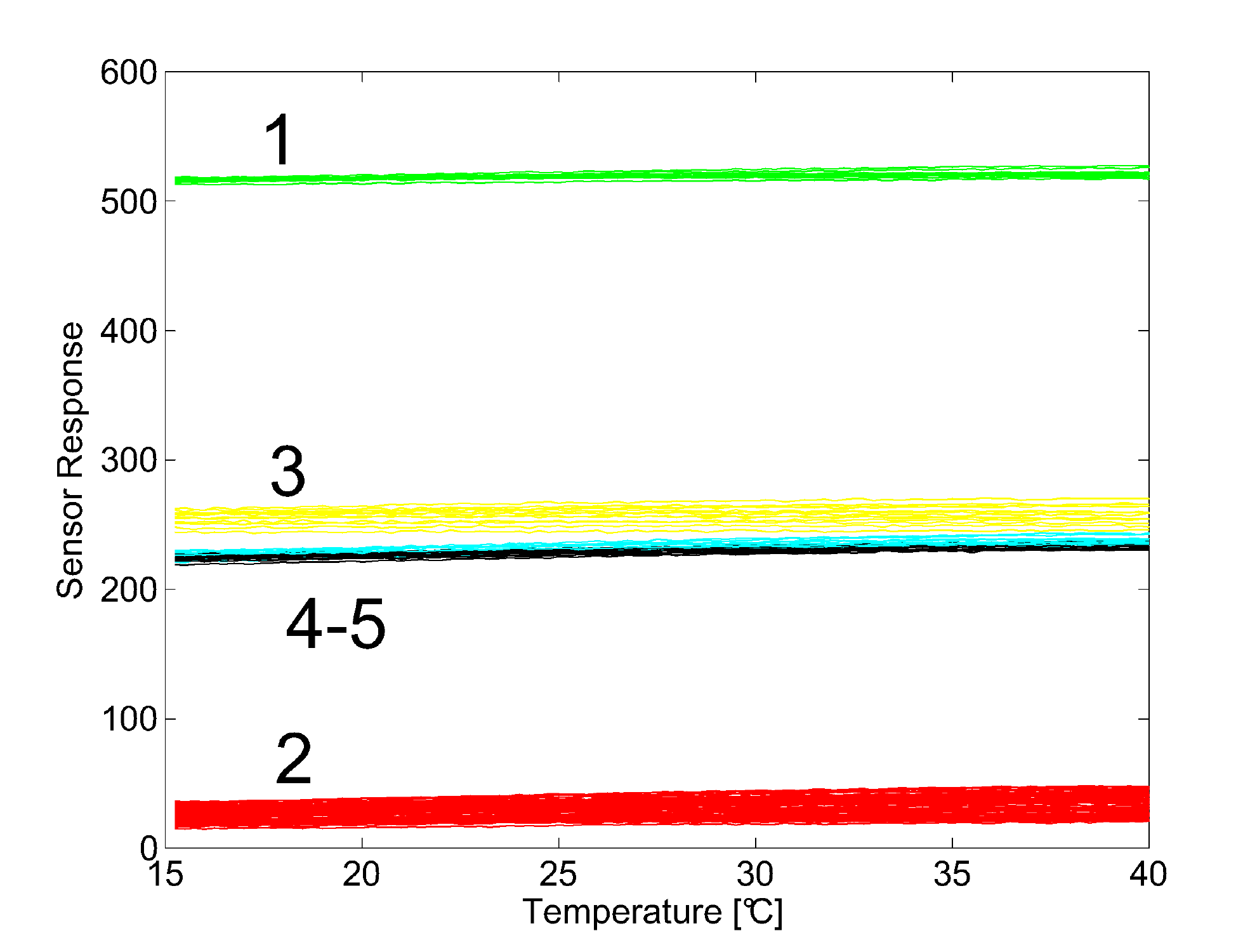}\label{1540comp}}\\
\caption{Temperature compensation, experiment 1. Different numbers represent different taxels as in Figure \ref{triangolo}, violet (thermal pads 6 and 7) represent the thermal sensors. In this case the temperature raised from 15 to 40 degrees. Top: the output of all sensors during the experiment. Bottom: result after the compensation. Notice that in the plot a different arbitrary offset was added to different sensors for better visualisation.}
\label{1540drift}
\end{figure}

\begin{figure}[t]
\centering
\subfigure[]{
\includegraphics[width=2.5in]{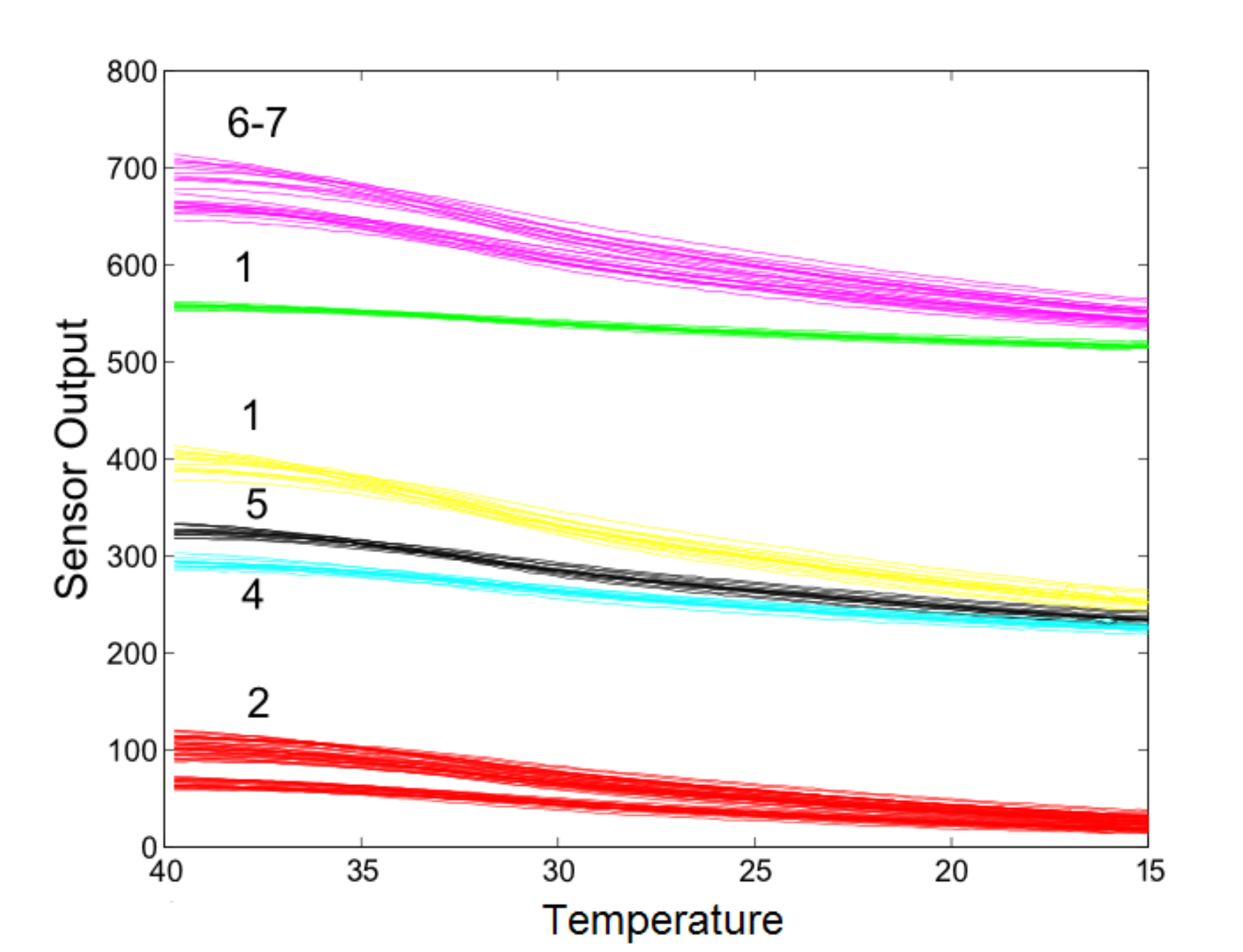}\label{4015}}
\hspace{0.25mm}
\subfigure[]{
\includegraphics[width=2.5in]{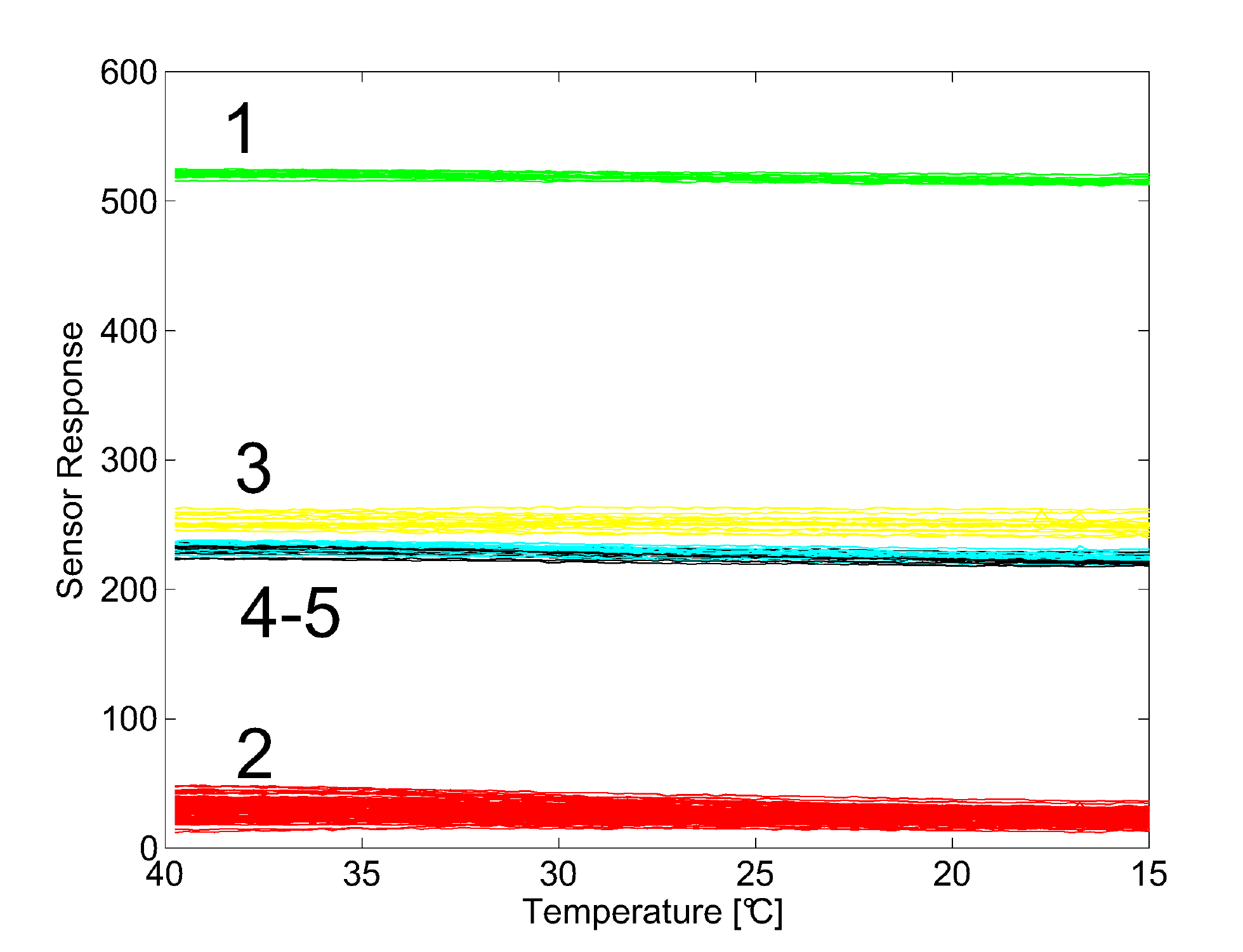}\label{4015comp}}\\
\caption{Temperature compensation, experiment 2. Different numbers represent  different sensors as in Figure \ref{triangolo}, violet (thermal pads 6 and 7) represent the thermal sensors. In this case the temperature lowered from 40 to 15 degrees. Top: the output of all sensors during the experiment. Bottom: result after the compensation.  Notice that in the plot a different arbitrary offset was added to different sensors for better visualisation.}
\label{1540drift}
\end{figure}

We performed experiments to asses the effectiveness of the compensation algorithm using a programmable oven and the 3D prototype (see Figure~\ref{upperarmsetup}). We performed two experiments: in the first experiment the temperature increased from $15^\circ C$ to $40^\circ C$, in the second experiment the temperature lowered from $40^\circ C$ to $15^\circ C$. The output of all taxels in a single triangle is reported in Figure~\ref{1540} and Figure~\ref{4015}. Different numbers were used to represent different taxels as illustrated in Figure~\ref{triangolo}. The curves identified by numbers 6 and 7 represent the 2 thermal sensors. In order to obtain the best result we determined different values of $K_{i}$ for each taxel (although Figure~\ref{1540} and Figure~\ref{4015} show that the thermal drift is similar for groups of taxels). 

The results in Figure \ref{1540comp}  and Figure \ref{4015comp} show that the compensation is effective in the range of temperatures we considered. We also determined that different triangles have similar thermal drift and that it is possible to use the same values of $K_{i}$ for different triangles without calibrating them individually.

\section{Conclusions}
\label{sec:conclusion}
In this paper we describe a new version of a modular, large scale capacitive tactile sensor for humanoid robots. The main improvement is that we replaced the elastomer used in the dielectric layer with a sandwich that includes a deformable 3D fabric, conductive Lycra and a protective layer. The production procedure of this sensor was therefore greatly simplified. In addition the new dielectric layer has better mechanical properties, it can be mounted easily without special glues and replaced if damaged. We characterised the response of the sensor in terms of repeatability, sensitivity, hysteresis and spatial resolution. We have shown that the sensor has good performance and in particular it has small hysteresis.  Finally, we demonstrated that the thermal sensors introduced in the FPCB can be effectively used to compensate drift induced by temperature variations. 

%


\ifCLASSOPTIONcaptionsoff
  \newpage
\fi



\bibliographystyle{IEEEtran}
\bibliography{soa}

\begin{thebibliography}{10}
\providecommand{\url}[1]{#1}
\csname url@samestyle\endcsname
\providecommand{\newblock}{\relax}
\providecommand{\bibinfo}[2]{#2}
\providecommand{\BIBentrySTDinterwordspacing}{\spaceskip=0pt\relax}
\providecommand{\BIBentryALTinterwordstretchfactor}{4}
\providecommand{\BIBentryALTinterwordspacing}{\spaceskip=\fontdimen2\font plus
\BIBentryALTinterwordstretchfactor\fontdimen3\font minus
  \fontdimen4\font\relax}
\providecommand{\BIBforeignlanguage}[2]{{%
\expandafter\ifx\csname l@#1\endcsname\relax
\typeout{** WARNING: IEEEtran.bst: No hyphenation pattern has been}%
\typeout{** loaded for the language `#1'. Using the pattern for}%
\typeout{** the default language instead.}%
\else
\language=\csname l@#1\endcsname
\fi
#2}}
\providecommand{\BIBdecl}{\relax}
\BIBdecl

\bibitem{Dahiya10}
R.~Dahiya, G.~Metta, M.~Valle, and G.~Sandini, ``Tactile sensing: From humans
  to humanoids,'' \emph{IEEE Transactions on Robotics}, vol.~26, no.~1, 2010.

\bibitem{Muhammad2011}
\BIBentryALTinterwordspacing
H.~Muhammad, C.~Oddo, L.~Beccai, C.~Recchiuto, C.~Anthony, M.~Adams,
  M.~Carrozza, D.~Hukins, and M.~Ward, ``Development of a bioinspired mems
  based capacitive tactile sensor for a robotic finger,'' \emph{Sensors and
  Actuators A: Physical}, vol. 165, no.~2, pp. 221 -- 229, 2011. [Online].
  Available:
  \url{http://www.sciencedirect.com/science/article/pii/S0924424710004838}
\BIBentrySTDinterwordspacing

\bibitem{kadowaki2009}
A.~Kadowaki, T.~Yoshikai, M.~Hayashi, and M.~Inaba, ``Development of soft
  sensor exterior embedded with multi-axis deformable tactile sensor system,''
  in \emph{Robot and Human Interactive Communication, 2009. RO-MAN 2009. The
  18th IEEE International Symposium on}.\hskip 1em plus 0.5em minus 0.4em\relax
  IEEE, 2009, pp. 1093--1098.

\bibitem{choi2005}
B.~Choi, H.~R. Choi, and S.~Kang, ``Development of tactile sensor for detecting
  contact force and slip,'' in \emph{Intelligent Robots and Systems, 2005.(IROS
  2005). 2005 IEEE/RSJ International Conference on}.\hskip 1em plus 0.5em minus
  0.4em\relax IEEE, 2005, pp. 2638--2643.

\bibitem{tajima2002}
R.~Tajima, S.~Kagami, M.~Inaba, and H.~Inoue, ``Development of soft and
  distributed tactile sensors and the application to a humanoid robot,''
  \emph{Advanced Robotics}, vol.~16, no.~4, pp. 381--397, 2002.

\bibitem{shear2007}
E.-S. Hwang, J.~hoon Seo, and Y.-J. Kim, ``A polymer-based flexible tactile
  sensor for both normal and shear load detections and its application for
  robotics,'' \emph{Microelectromechanical Systems, Journal of}, vol.~16,
  no.~3, pp. 556--563, June.

\bibitem{cmos2010}
Y.-C. Liu, C.-M. Sun, L.-Y. Lin, M.-H. Tsai, and W.~Fang, ``Development of a
  cmos-based capacitive tactile sensor with adjustable sensing range and
  sensitivity using polymer fill-in,'' \emph{Microelectromechanical Systems,
  Journal of}, vol.~20, no.~1, pp. 119--127, Feb.

\bibitem{mems2011}
M.~Ahmed, D.~Butler, and Z.-C. Butler, ``Mems relative pressure sensor on
  flexible substrate,'' in \emph{Sensors, 2011 IEEE}, Oct., pp. 460--463.

\bibitem{mannsfeld2010}
S.~C. Mannsfeld, B.~C. Tee, R.~M. Stoltenberg, C.~V.~H. Chen, S.~Barman, B.~V.
  Muir, A.~N. Sokolov, C.~Reese, and Z.~Bao, ``Highly sensitive flexible
  pressure sensors with microstructured rubber dielectric layers,''
  \emph{Nature materials}, vol.~9, no.~10, pp. 859--864, 2010.

\bibitem{Gray96}
B.~L. Gray and R.~S. Fearing, ``A surface micromachined microtactile sensor
  array,'' in \emph{Proc. Int. Conf. on Robotic Automation (ICRA)}, 1996.

\bibitem{Schmidt06}
P.~A. Schmidt, E.~Mael, and R.~P. Wurtz, ``A sensor for dynamic tactile
  information with applications in human-robot interaction \& object
  exploration,'' \emph{Robot. Autonomous Syst.}, vol.~54, pp. 1005--1014, 2006.

\bibitem{Miyazaki84}
S.~Miyazaki and A.~Ishida, ``Capacitive transducer for continuous measurement
  of vertical foot force,'' \emph{Med. Biol. Eng. Comput.}, vol.~22, p.
  309–316, 1984.

\bibitem{Schmitz08}
A.~Schmitz, M.~Maggiali, M.~Randazzo, L.~Natale, and G.~Metta, ``A prototype
  fingertip with high spatial resolution pressure sensing for the robot
  i{C}ub,'' in \emph{Proc. IEEE-RAS Int. Conf. on Humanoid Robots (Humanoids)},
  2008.

\bibitem{Schmitz2011}
A.~Schmitz, P.~Maiolino, M.~Maggiali, L.~Natale, G.~Cannata, and G.~Metta,
  ``Methods and technologies for the implementation of large-scale robot
  tactile sensors,'' \emph{Robotics, IEEE Transactions on}, vol.~27, no.~3, pp.
  389 --400, june 2011.

\bibitem{Inaba96}
M.~Inaba, Y.~Hoshino, K.~Nagasaka, T.~Ninomiya, S.~Kagami, and H.~Inoue, ``A
  full-body tactile sensor suit using electrically conductive fabric and
  strings,'' in \emph{Intelligent Robots and Systems '96, IROS 96, Proceedings
  of the 1996 IEEE/RSJ International Conference on}, vol.~2, nov 1996, pp. 450
  --457 vol.2.

\bibitem{Iwata09}
H.~Iwata, ``Design of human symbiotic robot {TWENDY-ONE},'' in \emph{Proc. IEEE
  Int. Conf. on Robotics and Automation (ICRA)}, 2009.

\bibitem{Ohmura06}
Y.~Ohmura, Y.~Kuniyoshi, and A.~Nagakubo, ``Conformable and scalable tactile
  sensor skin for curved surfaces,'' in \emph{Proc. IEEE Int. Conf. on Robotics
  and Automation (ICRA)}, 2006.

\bibitem{Mukai08}
T.~Mukai, M.~Onishi, S.~Hirano, and Z.~Luo, ``Development of the tactile sensor
  system of a human-interactive robot {``RI-MAN''},'' \emph{IEEE Transactions
  on Robotics}, vol.~24, no.~2, pp. 505--512, April 2008.

\bibitem{Asfour06}
T.~Asfour, K.~Regenstein, J.~Schroder, and R.~Dillmann, ``{ARMAR-III}: A
  humanoid platform for perception-action integration,'' in \emph{Proc. Int.
  Workshop on Human-Centered Robotic Systems}, 2006.

\bibitem{Mizuuchi06}
I.~Mizuuchi, T.~Yoshikai, T.~Nishino, and M.~Inaba, ``Development of
  musculoskeletal humanoid {K}otaro,'' in \emph{Proc. IEEE Int. Conf. on
  Robotics and Automation (ICRA)}, 2006.

\bibitem{Minato07}
T.~Minato, Y.~Yoshikawa, H.~Ishiguro, and M.~Asada, ``{CB2}: A child robot with
  biomimetic body for cognitive developmental robotics,'' in \emph{Proc.
  IEEE-RAS 7th Int. Conf. on Humanoid Robots (Humanoids)}, 2007.

\bibitem{Shimojo04}
M.~Shimojo, ``A tactile sensor sheet using pressure conductive rubber with
  electrical wires stitched method,'' \emph{IEEE Sensor Journal}, vol.~4,
  no.~5, 2004.

\bibitem{Mitt11}
P.~Mittendorfer and G.~Cheng, ``Humanoid multimodal tactile-sensing modules,''
  \emph{Robotics, IEEE Transactions on}, vol.~27, no.~3, pp. 401--410, june
  2011.

\bibitem{nielsen}
L.~Nielsen, \emph{Mechanical properties of polymers}.\hskip 1em plus 0.5em
  minus 0.4em\relax Van Nostrand Reinhold, 1962.

\bibitem{Lepora12}
N.~Lepora, U.~Martinez, H.~Barron, M.~Evans, G.~Metta, and T.~Prescott,
  ``Embodied hyperacuity from bayesian perception: Shape and position
  discrimination with an icub fingertip,'' in \emph{Proceedings of the IEEE/RSJ
  International Conference on Intelligent Robots and Systems}, 2012, pp.
  4638--4643.

\end{thebibliography}
\end{document}